\documentclass[11pt]{article}
\usepackage{latexsym,amssymb,amsmath,amsthm,graphics,graphicx,float,psfrag, epsfig, color, authblk}

\usepackage{xcolor}

\newtheorem{definition}{Definition}

\newtheorem{theorem}{Theorem}
\newtheorem{lemma}{Lemma}

\newcommand{\eps}{\varepsilon}
\newcommand{\R}{\mathbb{R}}

\newcommand{\<}{\langle}
\renewcommand{\>}{\rangle}

\newcommand{\erdosrenyi}{Erd\H{o}s-R\'{e}nyi\ }

\newcommand{\ti}{t_{i}}
\newcommand{\tj}{t_{j}}
\newcommand{\tk}{t_{k}}
\newcommand{\tl}{t_{\ell}}
\newcommand{\tnot}{t^{(0)}}
\newcommand{\pnot}{p^{(0)}}
\newcommand{\tbar}{\bar{t}}
\newcommand{\pbar}{\bar{p}}
\newcommand{\tpbar}{\zeta}

\newcommand{\toi}{t^{(0)}_{i}}
\newcommand{\poi}{p^{(0)}_{i}}

\newcommand{\vij}{v_{ij}}

\newcommand{\Eg}{E_g}
\newcommand{\Eb}{E_b}
\newcommand{\proj}{P}
\newcommand{\Pvijp}{\proj_{\vij^\perp}}

\renewcommand{\P}{\operatorname{\mathbb{P}}}
\newcommand{\tonum}[1]{t^{(0)}_{#1}}

\DeclareMathOperator{\Span}{span}

\newcommand{\vtildeij}{\tilde{v}_{ij}}
\newcommand{\Tbar}{\tbar}
\newcommand{\Tnot}{T^{(0)}}

\newcommand{\Pbar}{\pbar}
\newcommand{\Pnot}{P^{(0)}}

\newcommand{\zij}{z_{ij}}
\newcommand{\tij}{t_{ij}}
\newcommand{\tik}{t_{ik}}

\newcommand{\tjl}{t_{j\ell}}
\newcommand{\deltaij}{\delta_{ij}}
\newcommand{\deltatilde}{\tilde{\delta}}
\newcommand{\etaij}{\eta_{ij}}
\newcommand{\sij}{s_{ij}}
\newcommand{\lij}{\ell_{ij}}
\newcommand{\loij}{\ell^{(0)}_{ij}}
\newcommand{\toij}{t^{{(0)}}_{ij}}
\newcommand{\tokj}{t^{{(0)}}_{kj}}
\newcommand{\tokl}{t^{{(0)}}_{k\ell}}

\newcommand{\toil}{t^{{(0)}}_{i\ell}}
\newcommand{\toijperp}{t^{{(0)}\perp}_{ij}}
\newcommand{\toijhat}{\hat{t}^{(0)}_{ij}}

\newcommand{\tijhat}{\hat{t}_{ij}}

\newcommand{\poj}{p^{{(0)}}_{j}}

\newcommand{\muinf}{\mu_{\infty}}

\newcommand{\Vl}{V_\ell}
\newcommand{\Vs}{V_\text{s}}
\newcommand{\nl}{n_\ell}
\newcommand{\ns}{n_\text{s}}

\newcommand{\pj}{p_j}
\newcommand{\pl}{p_l}

\newcommand{\ahat}{\hat{a}}
\newcommand{\bhat}{\hat{b}}

\newcommand{\vhat}{\hat{v}}
\newcommand{\what}{\hat{w}}
\newcommand{\xhat}{\hat{x}}
\newcommand{\yhat}{\hat{y}}
\newcommand{\zhat}{\hat{z}}

\newcommand{\vtilde}{\tilde{v}}
\newcommand{\wtilde}{\tilde{w}}
\newcommand{\xtilde}{\tilde{x}}

\newcommand{\yxphat}{\hat{y}_{x^\perp}}

\newcommand{\Knn}{K_{\nl, \ns}}
\newcommand{\Kotilde}{\tilde{K}_0}

\newcommand{\ponum}[1]{p^{(0)}_{#1}}
\newcommand{\vijtilde}{\tilde{v}_{ij}}

\newcommand{\toitilde}{\tilde{t}^{(0)}_{i}}
\newcommand{\toktilde}{\tilde{t}^{(0)}_{k}}
\newcommand{\pojtilde}{\tilde{p}^{(0)}_{j}}
\newcommand{\poltilde}{\tilde{p}^{(0)}_{\ell}}
\newcommand{\Snot}{S^{(0)}}

\title{Exact simultaneous recovery of locations and structure from known orientations and corrupted point correspondences.}

\author{ Paul Hand*, Choongbum Lee and Vladislav Voroninski\\
  Department of Mathematics, Massachusetts Institute of Technology\\
  *Department of Computational and Applied Mathematics, Rice University
}

\begin{document}
\maketitle

\begin{abstract}
Let $t_1,\ldots,t_{n_l} \in \mathbb{R}^d$ and $p_1,\ldots,p_{n_s} \in \mathbb{R}^d$ and consider the bipartite location recovery problem: given a subset of pairwise direction observations $\{(t_i - p_j) / \|t_i - p_j\|_2\}_{i,j \in [\nl] \times [\ns]}$, where a constant fraction of these observations are arbitrarily corrupted, find $\{t_i\}_{i \in [\nl]}$ and $\{p_j\}_{j \in [\ns]}$ up to a global translation and scale. We study the recently introduced ShapeFit algorithm as a method for solving this bipartite location recovery problem.  In this case, ShapeFit consists of a simple convex program over $d(n_l + n_s)$ real variables. We prove that this program recovers a set of $n_l+n_s$ i.i.d. Gaussian locations exactly and with high probability if the observations are given by a bipartite \erdosrenyi graph, $d$ is large enough, and provided that at most a constant fraction of observations involving any particular location are adversarially corrupted. This recovery theorem is based on a set of deterministic conditions that we prove are sufficient for exact recovery. Finally, we propose a modified pipeline for the Structure for Motion problem, based on this bipartite location recovery problem. 

\end{abstract}

\section{Introduction}
Structure from Motion (SfM) is the task of recovering 3d structure from a collection of images taken from different vantage points \cite{StefanosBook}.  In the SfM problem, camera poses are represented by locations $\toi \in \mathbb{R}^3, i = 1\ldots \nl$ and rotation matrices $R_i \in SO(3), i =1 \ldots \nl$, where $R_i$ maps coordinates in the frame of the $i$th camera to the world frame. For a generic structure point $p \in \mathbb{R}^3$, there exists a unique point in each imaging plane given by perspective projection.  A pair of image points is said to correspond when they are both projections of the same point in 3d space.
 Given enough point correspondences between a pair of views, epipolar geometry yields the relative rotation and direction between  those views. Pairwise relative camera poses can then be used to estimate the individual poses $(\toi, R_i), i=1\ldots \nl$ up to a Euclidean transformation. Knowledge of camera poses and point correspondences allows one to estimate 3d structure via triangulation. Finally, the pose and structure estimates are used as initialization for bundle adjustment, which is the simultaneous nonlinear refinement of structure and camera poses. In summary, SfM typically consists of four steps: 1) identify point correspondences; 2) recover camera orientations and locations in global coordinates; 3) triangulate structure points using estimates of camera pose and correspondences; and 4) perform bundle adjustment.

A central difficulty of SfM is that point correspondences are prone to errors because they are found purely by local photometric information, which is subject to projective transformations from camera motion, specularities, occlusions, variable lighting conditions, shadows, and repetitive structures commonly found in manmade scenes. Thus, every step of the above SfM pipeline  needs to  tolerate highly corrupted input data. For the correspondence step, techniques such as Random Sampling Consensus (RANSAC) are used to reduce the number of outliers among candidate correspondences initially obtained by brute-force photometric matching. Unfortunately, even after applying RANSAC, outliers in point correspondences are generally unavoidable. 

Mathematically, once a set of correspondences has been established, the SfM problem can be formulated as the $d=3$ case of the following. Let $\Tnot$ be a collection of $\nl$ distinct vectors  $\tonum{1}, \ldots, \tonum{\nl} \in \R^d$, and let $\Pnot$ be a collection of $\ns$ distinct vectors  $\ponum{1}, \ldots, \ponum{\ns} \in \R^d$. Associated to locations $\Tnot$ is a set of orientations $R = \{R_i\}_{i \in [\nl]} \in SO(d)$.  The pairs $(\toi, R_i)$ represents poses from which observations  of the points $\poj$ are collected. Let $G(\nl,\ns, E)$ be a bipartite graph on $\nl + \ns$ vertices, where $E = \Eg \sqcup \Eb$, with $\Eb$ and $\Eg$ corresponding to pairwise direction observations that are respectively `corrupted' and `uncorrupted.'  That is, for each $ij \in E$, we are given a vector $\vij$, where
\begin{align*}
v_{ij} = \frac{R_i^t(\toi - \poj)}{\|R_i^t( \toi - \poj)\|_2} \text{ for } ij \in \Eg, \qquad \vij \in \mathbb{S}^{d-1} \text{ for } ij \in \Eb.
\end{align*}
An uncorrupted observation $\vij$ is exactly the direction of $R_i^t(\toi - \poj)$, and a corrupted observation is an arbitrary direction.  Consider the task of finding the unknown locations $\Tnot$ and structure points $\Pnot$, up to a global translation and scale, and the orientations $R$, up to a global rotation, without knowledge of the decomposition $E = \Eg \sqcup \Eb$, nor the nature of the corruptions. 

Estimating camera orientations $R_i$ from from corrupted relative rotations $R_i^t R_j$ is a tractable and relatively well-understood problem. For instance, a method based on Lie group averaging performs well in practice \cite{AMIT_13}, and a semidefinite program based on lifting and least unsquared deviations (LUD) has rigorous guarantees of exact recovery from corrupted relative rotations \cite{WangSingerSynchronization}.  Once camera orientations are estimated, one can use epipolar geometry to obtain a set of relative direction estimates of camera locations. These estimates are partially corrupted since they are computed from the initial point correspondences. Camera locations in a global reference frame can be estimated using the 1dSfM approach of \cite{1dsfm}, which screens for outliers based on inconsistencies in 1d projections; however, this approach is not robust to self-consistent outliers.  Alternatively, locations can be found by recent methods such as LUD \cite{AMIT} or the ShapeFit algorithm \cite{HLV2015}, which are both convex programs. It was proven in \cite{HLV2015} that ShapeFit recovers locations exactly from partially corrupted pairwise directions under broad technical assumptions. 

Having obtained an estimate of camera orientations and locations, one can recover an estimate of the 3d structure by triangulation, for instance by minimizing the quadratic reprojection error or maximizing a likelihood estimate. Bundle adjustment then proceeds by jointly optimizing this reprojection error or likelihood estimate with respect to camera poses and 3d structure. It is important to initialize bundle adjustment close to the global minimum, because it is  non-convex and susceptible to getting stuck in local minima.  

In this paper, we consider compressing two sub-steps of the pipeline --- camera location recovery and structure recovery by triangulation --- into one provably corruption-robust step based on the ShapeFit algorithm. Namely, once camera rotations are estimated, our approach uses the raw image coordinates of point correspondences to recover the camera locations and structure points simultaneously.  If a structure point $p_j$ is visible to a calibrated camera at location $t_i$, then its image coordinates under perspective projection provide a vector $\vijtilde $ that has the same direction as $R_i^t(t_i - p_j)$.  If the orientation $R_i$ is known and accurate, then the direction of $t_i - p_j$ is also known.  Equivalently,  if all the orientations $R_i$ are known, we can take each $R_i$ to be the identity without loss of generality. When a point correspondence is incorrect, the estimated direction of $t_i-p_j$ can of course be arbitrarily corrupted. We thus arrive at the following recovery problem.

 
With $T^{(0)}$ and $P^{(0)}$ defined as above, for each $ij \in E$, we are given a vector $\vij$, where
\begin{align}
\vij = \frac{\toi - \poj}{\bigl\| \toi - \poj \bigr\|_2} \text{ for } ij \in \Eg, \qquad \vij \in \mathbb{S}^{d-1} \text{ for } ij \in \Eb.  \label{location-recovery-measurements}
\end{align}
Thus, an uncorrupted observation $\vij$ is exactly the direction of $\toi - \poj$, and a corrupted observation is an arbitrary direction. The task is to find the unknown locations $\Tnot$, $\Pnot$ up to global translation and scale, without knowledge of the decomposition $E = \Eg \sqcup \Eb$, nor the nature of the corruptions.

To summarize, we propose the following modified pipeline for Structure from Motion: 1) establish point correspondences; 2) estimate global orientations of the cameras; 3) estimate the camera locations and structure points simultaneously; and 4) run bundle adjustment.

We will show that ShapeFit, a tractable convex program, can exactly solve the recovery problem in Step 3 under broad deterministic assumptions and under a random model. In \cite{HLV2015}, the present authors showed that ShapeFit recovers camera locations exactly from corrupted pairwise direction under suitable assumptions. The result in \cite{HLV2015} strongly relies on the existence of triangles in the graph of observations, whereas in our present setting, the underlying graphs are bipartite and necessarily do not contain triangles. In this bipartite setting, we will prove a deterministic recovery result for ShapeFit based on the presence of cycles of length 4. We also show that under a random Gaussian and Erdos-Renyi model, ShapeFit recovers structure and locations exactly from known orientations and corrupted correspondences with high probability in the high dimensional case. To the best of our knowledge, these are the first theoretical results guaranteeing exact location and structure recovery from corrupted correspondences and known orientations.

\subsection{ Problem formulation}
The location recovery problem is to recover a set of points in $\mathbb{R}^d$ from observations of pairwise directions between those points. Since relative direction observations are invariant under a global translation and scaling, one can at best hope to recover the locations $\Tnot = \{ \tonum{1},\ldots, \tonum{n}\}$ and structure points $\Pnot = \{ \ponum{1},\ldots, \ponum{n}\}$ up to such a transformation. That is, successful recovery from $\{v_{ij}\}_{(i,j) \in E}$ is finding two sets  of vectors ${\{\alpha(\toi + w)\}_{i \in [\nl]}}, {\{\alpha(\poj + w)\}_{j \in [\ns]}}$ for some $w \in \mathbb{R}^d$ and $\alpha >0$. We will say that two pairs of sets of vectors $(T,P) $ and $(\Tnot,\Pnot)$ are equal up to global translation and scale if there exists a vector $w$ and a scalar $\alpha >0 $ such that $t_i = \alpha(\toi + w)$ for all $i \in [\nl]$ and $p_j = \alpha(\poj + w)$ for all $j \in [\ns]$.  In this case, we will say that $(T,P)$ and $(\Tnot,\Pnot)$ have the same `shape,' and we will denote this property as $(T,P) \sim (\Tnot,\Pnot)$. The location recovery problem is then stated as:
\begin{alignat*}{2}
&\text{Given:} &\quad &G(\nl,\ns,E), \quad \{\vij\}_{ij \in E} \text{\ \  satisfying \eqref{location-recovery-measurements} }\\
&\text{Find:} && T = \{t_1, \ldots, t_{\nl} \} \in \mathbb{R}^{d \times \nl}, P = \{p_1, \ldots, p_{\ns} \} \in \mathbb{R}^{d \times \ns} \quad \text{such that} \quad (T,P) \sim (\Tnot,\Pnot) 
\end{alignat*}

For this problem to be information theoretically well-posed under arbitrary corruptions, the maximum number of corrupted observations affecting any particular location $t_i$ must be at most $\frac{\ns}{2}$.  Similarly, the maximum number affecting any particular structure point $p_j$ must be at most $\frac{\nl}{2}$.  Otherwise, suppose that for some location $\toi$ of structure point $\poj$, half of its associated observations $v_{ij}$ are consistent with $\toi$, and the other half are corrupted so as to be consistent with some arbitrary alternative location $w$. Distinguishing between $\toi$ and $w$ is then impossible in general. A similar argument follows for some structure point $\poj$. Formally, let $\deg_b(t_i)$ be the degree of location $t_i$ in the graph $G(\nl,\ns, \Eb)$ and let $\deg_b(p_j)$ be the degree of structure point $p_j$ in the graph $G(\nl,\ns, \Eb)$. Then, well-posedness under adversarial corruption requires that $\max_{i \in [\nl]} \deg_b(t_i) \leq \gamma \nl$ and $\max_{j \in [\ns]} \deg_b(v_j) \leq \gamma \ns$, for some $\gamma<1/2$,

Beyond the above necessary degree condition on $E_g$ for well-posedness of recovery, we do not assume anything about the nature of corruptions. That is, we work with adversarially chosen corrupted edges $E_b$ and arbitrary corruptions of observations associated to those edges. To solve the location recovery problem in this challenging setting, we utilize the convex program called ShapeFit \cite{HLV2015}:
\begin{align}
\min_{\substack{ \{t_i\}_{i \in [\nl]} \\ \{p_j\}_{j\in [\ns]}}} \ \sum_{ij \in E} \| \Pvijp (t_i - p_j) \|_2 \quad \text{ subject to } \quad \sum_{ij \in E} \langle t_i - p_j, \vij \rangle = 1, \ \ \  \sum_{i=1}^{\nl} t_i + \sum_{j=1}^{\ns} p_j = 0 \label{shapefit}
\end{align}
where $\Pvijp$ is the projector onto the orthogonal complement of the span of $\vij$.  

This convex program is a second order cone problem with $d(\nl + \ns)$ variables and two constraints.  Hence, the search space has dimension $d(\nl+\ns)-2$, which is minimal due to the $d(\nl + \ns)$ degrees of freedom in the locations $\{t_i\}$ and structure points $\{p_j\}$ and the two inherent degeneracies of translation and scale.

\subsection{Main result}

In this paper, we consider the model where the $\nl$ locations and $\ns$ structure points are i.i.d. Gaussian, and where pairwise direction observations are given according to an \erdosrenyi bipartite random graph.  We show that in a high-dimensional setting, ShapeFit \emph{exactly} recovers the locations and structure points with high probability, provided that $\nl$ and $\ns$ are sub-exponential in $d$, and provided that at most a fixed fraction of observations are \emph{adversarially} corrupted.

\begin{theorem} \label{thm:bipartite-random}
Let $N = \max(\nl,\ns), n = \min(\nl, \ns)$. Let $G(\Vl \cup \Vs,E)$ be drawn from a bipartite-\erdosrenyi graph with $p >0$. Take $ \tnot_1, \ldots \tnot_{\nl}, \pnot_1, \ldots, \pnot_{\ns} \sim \mathcal{N}(0, I_{d \times d})$ to be independent from each other and $G$. Then, there exist absolute constants $c,c_3, C>0$ such that for $\gamma = c_3 p^4$, if \\
$$\max\left(\frac{1}{c_3 p^4}, Cd, \frac{2 \log(eN)}{p}, \Omega(c_3 \log^2 N)\right) \leq  n \le N \leq e^{\frac{1}{8}c d }$$ 
and $d = \Omega(1)$, then there exists an event with probability at least  $1 - O( e^{-\Omega(\frac{1}{2}c_3^{-1/2} n^{1/2})} + e^{-\frac{1}{2} cd})$, on which the following holds:\\[1em]
For all subgraphs $\Eb$ satisfying $\max_{i\in [\nl]} \deg_b(\ti) \leq \gamma \ns$ and $\max_{j \in [\ns]} \deg_b(\pj) \leq \gamma \nl$ and all pairwise direction corruptions  $\vij \in \mathbb{S}^{d-1}$ for $ij \in \Eb$,  the convex program \eqref{shapefit} has a unique minimizer equal to $\left\{\alpha \{\toi -\tpbar \}_{i \in [\nl]}, \alpha \{\poi -\tpbar\}_{j \in [\ns]}  \right\}$ for some positive $\alpha$ and for $\tpbar = \frac{1}{\nl + \ns} \left(\sum_{i \in [\nl]} \toi + \sum_{j \in [\ns]} \poj \right)$. 
\end{theorem}

This probabilistic recovery theorem is based on a set of deterministic conditions that we prove are sufficient to guarantee exact recovery.  These conditions are satisfied with high probability in the model described above.  See Section \ref{sec-deterministic-theorem} for the deterministic conditions.

This recovery theorem is high-dimensional in the sense that the probability estimate and the exponential upper bound on $\nl + \ns$ are only meaningful for $d= \Omega(1)$.  Concentration of measure in high dimensions and the upper bound on $\nl + \ns$ ensure control over the angles and distances between random points. As a result, lower dimensional spaces are a more challenging regime for recovery.  

Numerical simulations empirically verify the main message of these recovery theorem: ShapeFit simultaneously recovers a set of locations and structure points exactly from corrupted direction observations, provided that up to a constant fraction of the observations at each location and structure point are corrupted. We present numerical studies in the physically relevant setting of $\mathbb{R}^3$, with an underlying random \erdosrenyi bipartite graph of observations. Further numerical simulations show that recovery is stable to the additional presence of noise on the uncorrupted measurements.  That is, locations and structure points are simultaneously recovered approximately under such conditions, with a favorable dependence of the estimation error on the measurement noise.

\subsection{Organization of the paper}
Section \ref{sec-notation} presents the notation used throughout the rest of the paper.   Section \ref{sec-proofs-high-d} presents the proof of Theorem \ref{thm:bipartite-random}.  Section \ref{sec-simulations} presents results from numerical simulations.

\subsection{Notation} \label{sec-notation}
Let $[k] = \{1, \ldots, k\}$.  Let $\Vl = [\nl]$ and $\Vs = [\ns]$.  Let $N = \max(\nl, \ns)$ and $n = \min(\nl, \ns)$. 
. Let $e_i$ be the $i$th standard basis element.  
For a bipartite graph $G(\Vl \cup \Vs, E)$, we write an arbitrary edge as an ordered pair $(i,j)$, where $i \in \Vl$ and $j \in \Vs$.  Let $\Knn$ be the complete bipartite graph on $\nl + \ns$ vertices.  A cycle of length 4 will be denoted as $C_4$.  Let $E(\Knn)$ be the set of edges in $\Knn$. 
Let  $\| \cdot \|_2$ be the standard $\ell_2$ norm on a vector. For any nonzero vector $v$, let $\hat{v} = v / \|v\|_2$.
For a subspace $W$, let $\proj_W$ be the orthogonal projector onto $W$. For a vector $v$, let $\proj_{v^\perp}$ be the orthogonal projector onto the orthogonal complement of the span of $\{v\}$.

Let $T$ denote the set $T = \{ \ti\}_{i \in \Vl }$, for $\ti \in \R^{d}$.  Let $P$ denote the set $P = \{\pj\}_{j \in \Vs}$, for $\pj \in \R^{d}$.  For $i \in \Vl, j \in \Vs$, define $\tij = \ti - \pj$
for all $i \in \Vl, j \in \Vs$.  For $i,k \in \Vl$, define $\tik = \ti - \tk$.  For $j,l \in \Vs$, define $\tjl = \pj - \pl$. 
Define $\bar{\zeta} = \frac{1}{\nl+\ns} \left( \sum_{i \in \Vl} \ti + \sum_{j \in \Vs} \pj  \right)$.  Define $\toij$, $\Tnot$, $\bar{\zeta}^{(0)}$, $\Pnot$, similarly. We define $\muinf = \max_{i\neq j} \|\toij\|_2$.  For a scalar $c$ and a set of vectors $X \subseteq \R^{d}$, let $c X = \{c x : x \in X\}$. 
 For a given $G=G(\Vl \cup \Vs,E)$ and $\{\vij\}_{ij \in E}$, where $\vij \in \R^{d}$ have unit norm, 
 let $R(T, P) = \sum_{ij \in E} \| \Pvijp \tij \|_2$.   Let $L(T,P) = \sum_{ij \in E} \< \tij, \vij \>$.  Let $\lij = \< \tij, \vij \>$, and similarly for $\loij$.  In this notation, ShapeFit is
\[
\min_{T,P} R(T,P) \quad \text{subject to} \quad  L(T,P) = 1, \quad \overline{\zeta}  = 0
\]
For vectors $v_1, \ldots, v_k$, let $S(v_1, \ldots, v_k) = \Span(v_1, \ldots, v_k)$ be the vector space spanned by these vectors.  
Given $\tij$ and $\toij$, define $\deltaij$, $\etaij$, and $\sij$ such that 
\[
\tij = (1 + \deltaij) \toij + \etaij \sij
\]
where $\sij$ is a unit vector orthogonal to $\toij$ and $\etaij = \| P_{\toijperp} \tij \|_2$.
Note that $\eta_{ij} \ge 0$.

\section{Proofs} \label{sec-proofs-high-d}

We will prove Theorem \ref{thm:bipartite-random} using the same general strategy as in \cite{HLV2015}.  Specifically, the proof of Theorem \ref{thm:bipartite-random} can be separated into two parts:  a recovery guarantee under a set of deterministic conditions, and a proof that the random model meets these conditions with high probability.  These sufficient deterministic conditions, roughly speaking, are (1) that the graph is connected and the nodes have tightly controlled degrees; (2) that the camera and structure locations are all distinct;  (3) that all pairwise distances between cameras and locations are within a constant factor of each other; (4) that any choice of two camera locations and two structure locations live in a three dimensional affine space; (5) that the camera and structure locations are `well-distributed' in a sense that we will make precise; and (6) that there are not too many corruptions affecting a single camera location or structure point.  Theorem \ref{thm:mainthm} in Section \ref{sec-deterministic-theorem} states these deterministic conditions formally.

As in \cite{HLV2015}, we will prove the deterministic recovery theorem directly, using several geometric properties concerning how deformations of a set of points induce rotations. 
Note that an infinitesimal rigid rotation of two points $\{t_i, t_j\}$ about their midpoint to ${\{t_i + h_i, t_j + h_j\}}$ is such that $h_i - h_j$ is orthogonal to $t_{ij} = t_i - t_j$.  We will abuse terminology and say that $\|P_{\tij^\perp} (h_i - h_j)\|$ is a measure of the rotation in a finite deformation $\{h_i, h_j\}$, and we say that $\langle h_i - h_j, \ti - \tj \rangle$ is the amount of stretching in that deformation. Using this terminology, the geometric properties we establish are:

\begin{itemize}
\item If a deformation stretches two adjacent sides of a $C_4$ at different rates, then that induces a rotation in some edge 
of the $C_4$ (Lemma \ref{lem:rigidity4}).  
\item If a deformation rotates one edge shared by many $C_4$s, then it induces a rotation over many of those $C_4$s, provided the opposite points of those triangles are `well-distributed' (Lemma \ref{lem:c4s}).
\item A deformation that rotates bad edges, must also rotate good edges (Lemma \ref{lem:badgood-bipartite}).
\item For any deformation, some fraction of the sum of all rotations must  affect the good edges (Lemma \ref{lem:transference-bipartite}).
\end{itemize}
By using these geometric properties, we show that all nonzero feasible deformations induce a
large amount of total rotation. Since some fraction of the total rotation must be on the good edges,  the objective must increase. 

The main technical difference between the present proof and the proof of \cite{HLV2015} is that the proof in $\cite{HLV2015}$ relies on the presence of many triangles in the graph of uncorrupted measurements.  Because of the bipartite structure of the present work, there are no triangles in the graph.  Hence, the technical novelty of the present proof is the establishment of the properties above when there are a sufficient number of $C_4$s in the graph of uncorrupted measurements.  

In Section \ref{sec-deterministic-theorem}, we present the deterministic recovery theorem.  In Section \ref{sec-unbalanced-motions}, we present and prove Lemma \ref{lem:rigidity4}.  In Section \ref{sec-c4-inequality}, we present and prove Lemmas \ref{lem:c4s}--\ref{lem:transference-bipartite}.  In Section \ref{sec-proof-mainthm}, we prove the deterministic recovery theorem.  In Section \ref{sec-properties-of-gaussians}, we prove that Gaussians satisfy several properties with high probability. In Section \ref{sec-gaussians-well-distributed}, we prove that Gaussians satisfy well-distributedness with high probability.  In Section \ref{sec-random-graph}, we prove that \erdosrenyi graphs are connected and have controlled degrees and codegrees with high probability.  Finally, in Section \ref{sec-proof-random-theorem}, we prove Theorem \ref{thm:bipartite-random}.


\subsection{Deterministic recovery theorem in high dimensions} \label{sec-deterministic-theorem}

To state the deterministic recovery theorem, we need two definitions.
The first definition captures the `regularity' of the measurement graph. A random bipartite graph can easily be seen to satisfy the conditions. Note that the definition does not depend on the vectors locations $\{t_i\}$ and $\{p_j\}$.

\begin{definition}
We say that a graph $G(\Vl \cup \Vs,E)$ is \emph{bipartite-$p$-typical} if it satisfies the following
properties:
\begin{enumerate}
  \setlength{\itemsep}{1pt} \setlength{\parskip}{0pt}
  \setlength{\parsep}{0pt}
	\item $G$ is connected,
	\item each vertex in $\Vl$ has degree between $\frac{1}{2}\ns p$ and $2\ns p$, and\\ each vertex in $\Vs$ has degree between $\frac{1}{2}\nl p$ and $2 \nl p$.
	\item each pair of vertices in $\Vl$ has codegree between $\frac{1}{2}\ns p^2$ and $2 \ns p^2$, where the codegree  of $j,l \in \Vl = |\{i \mid ij \in E(G), il \in E(G) \} |$.  Each pair of vertices in $\Vs$ has codegree between $\frac{1}{2}\nl p^2$ and $2 \nl p^2$.
\end{enumerate}
\end{definition}

The next definition captures how `well-distributed' the location points $\{t_i\}$ and $\{p_j\}$ are in $\R^d$.  

\begin{definition} \ 
\begin{itemize}
  \setlength{\itemsep}{1pt} \setlength{\parskip}{0pt}
  \setlength{\parsep}{0pt}
\item[(i)] Let $S = \{(t_k, p_k)\}_{k=1\ldots m} \subset \R^d \times \R^d$. Let $x,y \in \R^d$.
We say that $S$ is $c$-well-distributed with respect to $(x,y)$ if the following holds for all $h \in \R^d$:
\[
	\sum_{(t,p) \in S}\|\proj_{\Span\{p-x,t-p, y-t\}^{\perp}}(h)\|_2 \ge c|S|\cdot\|\proj_{(x-y)^{\perp}}(h)\|_2.
\]
\item[(ii)] Let $T = \{\ti\}_{i \in \Vl}$ and $P = \{\pj\}_{j \in \Vs}$.  We say that $(T,P)$ is c-well-distributed along $G$ if for all $i \in \Vl, j \in \Vs$, the set $S_{ij} = \{(t_k, p_\ell)\,:\, i\ell \in E(G), k\ell \in E(G), kj \in E(G), k \neq i, \ell \neq j \}$ is $c$-well-distributed with respect to $(t_i, p_j)$.
\end{itemize}
\end{definition}

We now state sufficient deterministic recovery conditions on the graph $G$, the subgraph $\Eb$ corresponding to corrupted observations, and the locations $\Tnot$ and $\Pnot$.

\begin{theorem} \label{thm:mainthm}
Suppose $\Tnot, \Pnot, \Eb, G$ satisfy the conditions
\begin{enumerate}
  \setlength{\itemsep}{1pt} \setlength{\parskip}{0pt}
  \setlength{\parsep}{0pt}
	\item The underlying graph $G$ is bipartite-$p$-typical,
	\item All vectors in $\Tnot$, $\Pnot$ and $\Tnot \cup \Pnot$ are distinct, respectively.
	\item For all $i,k \in \Vl$ and $j, \ell \in \Vs$, we have $c_{0}\|t_{k\ell}^{(0)}\|_2\le\|t_{ij}^{(0)}\|_2$,
	\item For all $i,k \in \Vl, j,\ell \in \Vs$ such that $k \neq i, j \neq \ell$, we have \\
	$\min \Bigl(\|\proj_{\Span (\tokj,\toil)^{\perp}} \toij \|_2, 
	 \|\proj_{\Span (\tokl,\toil)^{\perp}} \toij \|_2 \Bigr) / \|\toij \|_2 \geq \beta$
	\item The pair $(\Tnot, \Pnot)$ is $c_{1}$-well-distributed
along $G$,
	\item Each vertex in $\Vl$ (resp.  $\Vs$) has at most $\eps \ns$ (resp.  $\eps \nl$) incident edges in $\Eb$.
\end{enumerate}
for constants $0< p, c_0, \beta,  c_{1}, \eps \leq 1$.  If $\eps \leq \frac{\beta c_0 c_1^2 p^4}{384\cdot204\cdot64}$ and $\nl, \ns > max(64, \frac{8}{p^2})$, then $L(\Tnot, \Pnot)\neq 0$ and $(\Tnot, \Pnot) / L(\Tnot, \Pnot)$ is the unique optimizer of ShapeFit.
\end{theorem}

Before we prove the theorem, we establish that $L(\Tnot, \Pnot)\neq 0$ when $\eps$ is small enough.  This property guarantees that some scaling of $(\Tnot,\Pnot)$ is feasible and occurs, roughly speaking, when $|\Eb| < |\Eg|$.

\begin{lemma} \label{lem:T0-feasible-bipartite} 
If $\varepsilon < \frac{c_0 p}{4}$, then $L(\Tnot, \Pnot) \neq 0$.
\end{lemma}
\begin{proof}
Since $v_{ij} = \toijhat$ for all $ij \in E_g$, we have
\[
L(\Tnot,\Pnot) = \sum_{ij \in E(G)} \langle \toij, v_{ij} \rangle \geq \sum_{ij \in E_g} \| \toij\|_2 - \sum_{ij \in E_b} \| \toij \|_2.
\]
By Condition 3,  $c_0 \muinf |E_g| \leq \sum_{ij \in E_g} \| \toij\|_2$ and $\muinf |E_b| \geq  \sum_{ij\in E_b} \| \toij\|_2$.  Thus it suffices to prove that $c_0 |E_g| > |E_b|$. 
As $\varepsilon < \frac{p}{4}$, Condition 1 and 6 gives $|E_g| \geq \frac{1}{2} \nl \ns p - \eps \nl \ns \geq \frac{1}{4}\nl \ns p$. Since $|E_b| \leq \eps \nl \ns$, if $\eps < \frac{c_0 p}{4}$, then we have $c_0 |E_g| > |E_b|$. 
\end{proof}

\subsection{Unbalanced parallel motions induce rotation} \label{sec-unbalanced-motions}

The following lemma concerns geometric properties of deformations of a set of points.  Specifically it shows that if four points are deformed in a way that differentially scales the lengths of two edges, then it necessarily induces a rotation somewhere in a $C_4$ containing those points.


\begin{lemma} \label{lem:rigidity4}
Let $d\ge3$. Let $t_{1},t_{2},t_{3},t_{4} \in \R^d$ be distinct.  Let $t_{ij} = \ti - \tj$ and $\tijhat = \frac{\tij}{\| \tij \|}$.  Let $v_{1},v_{2},v_{3},v_{4}\in\mathbb{R}^{d}$ and $\alpha \in \R$.   Let $\{\deltatilde_{i(i+1)}\}$ be such that  ${\langle v_{i}-v_{i+1}-\alpha t_{i(i+1)},\hat{t}_{i(i+1)}\rangle=}$ $\deltatilde_{i(i+1)}\|t_{i(i+1)}\|_2$ for each $i\in[4]$, where index summation is modulo 4.
\begin{align*}
(i) \quad &\sum_{i\in[4]} 
	\|\proj_{t_{i(i+1)}^{\perp}}(v_{i}-v_{i+1})\|_2
	\ge \left\| \proj_{\Span(t_{23},t_{41})^{\perp}}t_{12}\right\|_2 \left|\deltatilde_{12}-\deltatilde_{34}\right| \\
(ii) \quad &\sum_{i\in[4]} 
	\|\proj_{t_{i(i+1)}^{\perp}}(v_{i}-v_{i+1})\|_2
	\ge  \left\| \proj_{\Span(t_{34},t_{41})^{\perp}} t_{12}\right\|_2 \left|\deltatilde_{12}-\deltatilde_{23}\right|.
\end{align*}
\end{lemma}
\begin{proof}
The given condition implies $\proj_{t_{i(i+1)}^{\perp}}(v_{i}-v_{i+1})=v_{i}-v_{i+1}-(\alpha+\deltatilde_{i(i+1)})t_{i(i+1)}$
for each $i \in [4]$. Therefore,
\begin{eqnarray}
\sum_{i \in [4]}\|\proj_{t_{i(i+1)}^{\perp}}(v_{i}-v_{i+1})\|_2 & = & \sum_{i \in [4]}\left\| v_{i}-v_{i+1}-\left(\alpha+\deltatilde_{i(i+1)}\right)t_{i(i+1)}\right\|_2 \nonumber \\
 & \ge & \left\| \sum_{i \in [4] }v_{i}-v_{i+1}-\left(\alpha+\deltatilde_{i(i+1)}\right)t_{i(i+1)}\right\|_2 \nonumber \\
 & = & \|\deltatilde_{12}t_{12}+\deltatilde_{23}t_{23}+\deltatilde_{34}t_{34}+\deltatilde_{41}t_{41}\|_2. \label{eq:rigidity}
\end{eqnarray}

\noindent (i) Since $\deltatilde_{34}(t_{12}+t_{23}+t_{34}+t_{41})=0$, the right-hand-side of
\eqref{eq:rigidity} equals $\|(\deltatilde_{12}-\deltatilde_{34})t_{12}+(\deltatilde_{23}-\deltatilde_{34})t_{23}+(\deltatilde_{41}-\deltatilde_{34})t_{41}\|_2$.
The conclusion follows since
\begin{eqnarray*}
\left\| (\deltatilde_{12}-\deltatilde_{34})t_{12}+(\deltatilde_{23}-\deltatilde_{34})t_{23}+(\deltatilde_{41}-\deltatilde_{34})t_{41}\right\|_2  & \ge & \min_{s,s'\in\mathbb{R}}\|(\deltatilde_{12}-\deltatilde_{34})t_{12}-st_{23}-s't_{41}\|_2\\
 & = & \left\| \proj_{\Span(t_{23},t_{41})^{\perp}}(\deltatilde_{12}-\deltatilde_{34})t_{12}\right\|_2.
\end{eqnarray*}

\noindent (ii) Since $\deltatilde_{23}(t_{12}+t_{23}+t_{34}+t_{41})=0$, the right-hand-side
of \eqref{eq:rigidity} equals $\|(\deltatilde_{12}-\deltatilde_{23})t_{12}+(\deltatilde_{34}-\deltatilde_{23})t_{34}+(\deltatilde_{41}-\deltatilde_{23})t_{41}\|_2$.
The conclusion follows since
\begin{eqnarray*}
\left\| (\deltatilde_{12}-\deltatilde_{23})t_{12}+(\deltatilde_{34}-\deltatilde_{23})t_{34}+(\deltatilde_{41}-\deltatilde_{23})t_{41}\right\|_2  & \ge & \min_{s,s'\in\mathbb{R}}\|(\deltatilde_{12}-\deltatilde_{23})t_{12}-st_{34}-s't_{41}\|_2\\
 & = & \left\| \proj_{\Span(t_{34},t_{41})^{\perp}}(\deltatilde_{12}-\deltatilde_{23})t_{12}\right\|_2.\qedhere
\end{eqnarray*}
\end{proof}

\subsection{$C_4$s inequality and rotation propagation} \label{sec-c4-inequality}

The following lemma is a generalization of the triangle inequality in a context of the rotational part of structure deformations.

\begin{lemma}[$C_4$s Inequality] \label{lem:c4s}
Let $d\ge4$; $x,y \in \mathbb{R}^{d}$.  Let $S=\{(t_1, p_1), \cdots, (t_k, p_k)\} \subset \R^d \times \R^d$.  If $S$  is $c$-well-distributed
with respect to $(x,y)$, then for all vectors $h_{x},h_{y},h_{t_1},\cdots,h_{t_k},h_{p_1},\cdots,h_{p_k}\in\mathbb{R}^{d}$
and sets $X\subseteq[k]$, we have
\[
\sum_{i\in[k]\setminus X}\|\proj_{(x-p_{i})^{\perp}}(h_{x}-h_{p_i})\|_2 + 
\|\proj_{(p_{i}-t_i)^{\perp}}(h_{p_i}-h_{t_i})\|_2 + \|\proj_{(t_{i}-y)^{\perp}}(h_{t_i}-h_{y})\|_2 \ge(ck-|X|)\cdot\|\proj_{(x-y)^{\perp}}(h_{x}-h_{y})\|_2.
\]
\end{lemma}
\begin{proof}
For each $i\in[k]$, define $W_i = \Span(x - p_i, p_i - t_i, t_i - y)$.
Define $P$ as the projection map to the space of vectors orthogonal
to $x-y$, and define $P_{i}$ for each $i\in[k]$  as the projection
map to $W_{i}^{\perp}$. Since $(x-p_{i})^{\perp}\supseteq W_{i}^{\perp}$, $(p_i - t_i)^{\perp}\supseteq W_{i}^{\perp}$, and $(t_i - y)^{\perp}\supseteq W_{i}^{\perp}$,
it follows that 
\begin{eqnarray*}
 &  &\sum_{i\in[k]\setminus X}\|\proj_{(x-p_{i})^{\perp}}(h_{x}-h_{p_i})\|_2 + 
\|\proj_{(p_{i}-t_i)^{\perp}}(h_{p_i}-h_{t_i})\|_2 + \|\proj_{(t_{i}-y)^{\perp}}(h_{t_i}-h_{y})\|_2 \\
 & \ge & \sum_{i\in[k]\setminus X}\|P_{i}(h_{x}-h_{p_i})\|_2 +\|P_{i}(h_{p_i}-h_{t_i})\|_2 +\|P_{i}(h_{t_i}-h_{y})\|_2 \ge\sum_{i\in[k]\setminus X}\|P_{i}(h_{x}-h_{y})\|_2.
\end{eqnarray*}
Since $\{(t_1, p_1), \cdots, (t_k, p_k)\}$ are well-distributed with respect to $(x,y)$,
we have
\begin{equation}
\sum_{i\in[k]}\|P_{i}(h_{x}-h_{y})\|_2 \ge ck\cdot\|P(h_{x}-h_{y})\|_2.\label{eq:proj_triangles}
\end{equation}
Since $(x-y)^\perp \supseteq W_i^\perp$, we have $\|P_{i}(h_{x}-h_{y})\|_2\le\|P(h_{x}-h_{y})\|_2$ for all $i$. Hence
\[
\sum_{i\in[k]\setminus X}\|P_{i}(h_{x}-h_{y})\|_2 \ge(ck-|X|)\cdot\|P(h_{x}-h_{y})\|_2,
\]
proving the lemma.
\end{proof}

The proof of Theorem \ref{thm:mainthm} will rely on the following two lemmas, which state that rotational motions on some parts of the graph bound rotational motions on other parts. The following lemma relates the rotational motions on bad edges to the rotational motions on good edges.  Recall the notation $\tij = (1 + \deltaij) \toij + \etaij \sij$ where $\sij$ is a unit vector orthogonal to $\toij$ and $\etaij = \| P_{\toijperp} \tij \|_2$.

\begin{lemma} \label{lem:badgood-bipartite} 
Fix $T, P$. If $\eps \leq \frac{c_1 p^3}{48}$ and $p \geq \sqrt{\frac{8}{n}}$, then $\sum_{ij\in E_{g}}\eta_{ij}\ge\frac{c_{1}p^{3}}{48\eps}\sum_{ij\in E_{b}}\eta_{ij}$. 
\end{lemma}
\begin{proof}
Let $i \in \Vl, j \in \Vs$.  Note that Condition 1 implies $|\{(k,\ell) \mid k\neq i; \ell \neq j; i\ell,k\ell,kj\in E(G) \}|\geq (\frac{1}{2} \ns p -1) (\frac{1}{2} \nl p^2 - 1) \geq \frac{1}{8} \nl \ns p^3$ if $p \geq \sqrt{\frac{8}{n}}$. By Condition 6, the number of pairs $(k,\ell) \in \Vl \times \Vs$ such that at least one of the edges $i\ell, k\ell, kj$ are in $E_b$ can be counted by considering the case when $i\ell \in E_b$ (at most $(\varepsilon \ns) \nl$ pairs), $kj \in E_b$ (at most $(\varepsilon \nl) \ns$ pairs), and $k\ell \in E_b$ (at most $\varepsilon \ns \nl$ pairs). Hence in total, there are at most $3\varepsilon \ns \nl$ such pairs.
  By Lemma \ref{lem:c4s}, the $c_1$-well-distributedness of $(\Tnot,\Pnot)$ along $G$, and the assumption that $\eps \leq  \frac{c_1 p^3}{48}$, we have 
\[
	\sum_{\substack{k \in \Vl, \ell \in \Vs\\k\neq i,\ell \neq j\\i\ell, k\ell, kj\in E_{g}}}(\eta_{i\ell}+\eta_{k\ell} + \eta_{kj})
	\ge\left(c_{1}\cdot\frac{1}{8}\nl \ns p^3- 3 \eps \nl \ns\right)\cdot\eta_{ij}
	\ge\frac{c_{1}}{16}\nl \ns p^3\cdot\eta_{ij}.
\]
Therefore, if we sum the inequality above for all bad edges $ij\in E_{b}$, then
\[
	\sum_{ij\in E_{b}}
	\sum_{\substack{k \in \Vl, \ell \in \Vs\\k\neq i,\ell \neq j\\il, k\ell, kj\in E_{g}}}(\eta_{i\ell}+\eta_{k\ell} + \eta_{kj})
\ge\frac{c_{1}}{16}\nl \ns p^3\cdot\sum_{ij\in E_{b}}\eta_{ij}.
\]

For fixed $k\ell\in E_{g}$, the left-hand-side may sum $\eta_{k\ell}$
as many times as the number of $C_4$s in $E(G)$ that contain $k\ell$ and exactly one bad edge.  This is the same as the number of $C_4$s whose edge opposite $k\ell$ is bad, plus the number of $C_4$s whose edge adjacent to $\ell$ is bad, plus the number of $C_4$s whose edge adjacent to $k$ is bad.   In each case, there are at most $\eps \nl \ns$ such $C_4$s.
Hence, the left-hand-side of above is at most 
\[
	\sum_{ij\in E_{b}}
	\sum_{\substack{k \in \Vl, \ell \in \Vs\\k\neq i,\ell \neq j\\i\ell, k\ell, kj\in E_{g}}}(\eta_{i\ell}+\eta_{k\ell} + \eta_{kj})
	\le 3 \eps \nl \ns\cdot\sum_{ij\in E_{g}}\eta_{ij}.
\]
Therefore by combining the two inequalities above, we obtain 
\[
	\sum_{ij\in E_{b}}\eta_{ij}
	\le\frac{48\eps}{c_{1}p^{3}}\sum_{ij\in E_{g}}\eta_{ij}. 
	\qedhere
\]
\end{proof}

The following lemma relates the rotational motions over the good graph $\Eg$ to rotational motions over the complete bipartite graph $\Knn$.

\begin{lemma} \label{lem:transference-bipartite} 
Fix $T, P$. If $\eps \leq \frac{c_1 p^3}{48}$ and $p \geq \sqrt{\frac{8}{n}}$, then $\sum_{ij\in E_{g}}\eta_{ij}\ge\frac{c_{1}p}{192}\sum_{ij\in E(\Knn)}\eta_{ij}$.
\end{lemma}
\begin{proof}
Let $i \in \Vl, j \in \Vs$.  Note that Condition 1 implies $|\{(k,\ell) \mid k\neq i; \ell \neq j; i\ell,k\ell,kj\in E(G) \}|\geq (\frac{1}{2} \ns p -1) (\frac{1}{2} \nl p^2 - 1) \geq \frac{1}{8} \nl \ns p^3$ if $p \geq \sqrt{\frac{8}{n}}$.  Similarly as in Lemma~\ref{lem:badgood-bipartite}, Condition 6 implies that
the number of pairs $(k,\ell) \in \Vl \times \Vs$ such that at least one of the edges $i\ell, k\ell, kj$ are in $E_b$ is at most $3\eps \nl \ns$.
By Lemma \ref{lem:c4s}, the $c_1$-well-distributedness of $(\Tnot,\Pnot)$ along $G$, and the assumption that $\eps \leq  \frac{c_1 p^3}{48}$, we have 
\[
	\sum_{\substack{k \in \Vl, \ell \in \Vs\\k\neq i,\ell \neq j\\i\ell, k\ell, kj\in E_{g}}}(\eta_{i\ell}+\eta_{k\ell} + \eta_{kj})
	\ge\left(c_{1}\cdot\frac{1}{8}\nl \ns p^3- 3 \eps \nl \ns\right)\cdot\eta_{ij}
	\ge\frac{c_{1}}{16}\nl \ns p^3\cdot\eta_{ij}.
\]
Therefore, if we sum the inequality above for all $i\in \Vl, j \in \Vs$, or equivalently over all $ij \in E(\Knn)$, then
\[
	\sum_{ij \in E(\Knn)}
	\sum_{\substack{k \in \Vl, \ell \in \Vs\\k\neq i,\ell \neq j\\i\ell, k\ell, kj\in E_{g}}}(\eta_{i\ell}+\eta_{k\ell} + \eta_{kj})
\ge\frac{c_{1}}{16}\nl \ns p^3\cdot\sum_{ij\in E(\Knn)}\eta_{ij}.
\]

For fixed $k\ell\in E_{g}$, the left-hand-side may sum $\eta_{k\ell}$
as many times as the number of paths of length $3$ in $G$ that contain $k\ell$.  Each path of length $3$ can be thought of as an edge originating from $\Vl$, an edge in the middle, and an edge terminating in $\Vs$.  The total number of paths of length $3$ in $G$ containing $k\ell$ equals the number which have $k\ell$ as the middle edge, plus the number with $k\ell$ as the edge originating from $\Vl$, plus the number with $k\ell$ as the edge terminating in $\Vs$.   In each of these cases, Condition 1 ensures that there are at most $ 4 p^2 \nl \ns$ such paths of length $3$.  Hence, the term $\eta_{k\ell}$ appears at most $12 p^2 \nl \ns$ times.
Hence, the left-hand-side of above is at most 
\[
	\sum_{ij\in E(\Knn)}
	\sum_{\substack{k \in \Vl, \ell \in \Vs\\k\neq i,\ell \neq j\\i\ell, k\ell, kj\in E_{g}}}(\eta_{i\ell}+\eta_{k\ell} + \eta_{kj})
	\le 12 p^2 \nl \ns\cdot\sum_{ij\in E_{g}}\eta_{ij}.
\]
Therefore by combining the two inequalities above, we obtain 
\[
	\sum_{ij\in E(\Knn)}\eta_{ij}
	\le\frac{12\cdot 16}{c_{1}p}\sum_{ij\in E_{g}}\eta_{ij}. 
	\qedhere
\]

\end{proof}

\subsection{Proof of Theorem \ref{thm:mainthm}} \label{sec-proof-mainthm}

We now prove the deterministic recovery theorem.

\begin{proof}[Proof of Theorem \ref{thm:mainthm}]

By Lemma \ref{lem:T0-feasible-bipartite} and the fact that Conditions 1--6 are invariant under global translation and nonzero scaling, we can take $\bar{\zeta}^{(0)} =0$ and $L(\Tnot, \Pnot) = 1$ without loss of generality.  The variable $\muinf =  \max_{i\neq j} \|\toij\|_2$ is to be understood accordingly.

We will directly prove that $R(T, P) > R(\Tnot, \Pnot)$ for all $(T, P) \neq (\Tnot, \Pnot)$ such that $L(T, P)=1$ and $\Tbar + \Pbar = 0$. Consider an arbitrary feasible $T,P$ and recall the notation $\tij = (1 + \deltaij) \toij + \etaij \sij$ where $\sij$ is a unit vector orthogonal to $\toij$ and $\etaij = \| P_{\toijperp} \tij \|_2$. Since $v_{ij} =\toijhat$ holds for all $ij \in E_g$, a useful lower bound for the objective $R(T,P)$ is given by 
\begin{eqnarray}
	R(T,P)
	=\sum_{ij \in E(G)}\|\proj_{v_{ij}^{\perp}}t_{ij}\| _2
	& = & \sum_{ij\in E_{g}}\eta_{ij}+\sum_{ij\in E_{b}}\|\proj_{v_{ij}^{\perp}}t_{ij}\|_2\nonumber \\
	& \ge & \sum_{ij\in E_{g}}\eta_{ij}+\sum_{ij\in E_{b}}\left(\|\proj_{v_{ij}^{\perp}}t_{ij}^{(0)}\|_2-|\delta_{ij}|\|t_{ij}^{(0)}\|_2-\eta_{ij}\right)\nonumber \\
	& \ge & R(\Tnot, \Pnot)+\sum_{ij\in E_{g}}\eta_{ij}-\sum_{ij\in E_{b}}(|\delta_{ij}|\|t_{ij}^{(0)}\|_2+\eta_{ij}).\label{eq:gain3-bipartite}
\end{eqnarray}

Suppose that $\sum_{ij\in E_{b}}|\delta_{ij}|\|t_{ij}^{(0)}\|_2<\sum_{ij\in E_{b}}\eta_{ij}$.
Since Lemma \ref{lem:badgood-bipartite} for $\varepsilon\le\frac{c_{1}p^3}{96}$
implies $\sum_{ij\in E_{b}}\eta_{ij}\le\frac{1}{2}\sum_{ij\in E_{g}}\eta_{ij}$,
by (\ref{eq:gain3-bipartite}), we have
\begin{eqnarray*}
	R(T,P) & \ge & R(\Tnot, \Pnot)+\sum_{ij\in E_{g}}\eta_{ij}-\sum_{ij\in E_{b}}(|\delta_{ij}|\|t_{ij}^{(0)}\|_2+\eta_{ij})\\
	& > & R(\Tnot, \Pnot)+\sum_{ij\in E_{g}}\eta_{ij}-\sum_{ij\in E_{b}}2\eta_{ij}\ge R(\Tnot, \Pnot).
\end{eqnarray*}
Hence we may assume 
\begin{equation}
	\sum_{ij\in E_{b}}|\delta_{ij}|\|t_{ij}^{(0)}\|_2\ge\sum_{ij\in E_{b}}\eta_{ij}.\label{eq:badtwist-bipartite}
\end{equation}

In the case $|\Eb| \neq 0$, define $\overline{\delta}=\frac{1}{|E_{b}|}\sum_{ij\in E_{b}}|\delta_{ij}|$
as the average `relative parallel motion' on the bad edges. For a pair of vertex-disjoint edges $ij,k\ell\in E(\Knn)$, define $\eta(ij,k\ell)=\eta_{ij} + \eta_{kj} + \eta_{k\ell} + \eta_{i\ell}$,
\\

\noindent \textbf{Case 0}. $|\Eb| = 0$ or $\bar{\delta} = 0$.
\medskip

Note that $\bar{\delta}=0$ implies $\delta_{ij} =0$ for all $ij \in E_b$, which by \eqref{eq:badtwist-bipartite} implies
$\eta_{ij} =0$ for all $ij \in E_b$.  Therefore by \eqref{eq:gain3-bipartite}, we have
\[
	R(T, P) \geq R(\Tnot, \Pnot) + \sum_{ij \in E_g} \eta_{ij}.
\]
If $\sum_{ij \in E_g} \eta_{ij} > 0$, then we have $R(T, P) > R(\Tnot, \Pnot)$. Thus 
we may assume that $\eta_{ij} = 0$ for all $ij \in E_g$. In this case, we will show that $T = \Tnot$ and $P = \Pnot$.

By Lemma~\ref{lem:transference-bipartite}, if $\eps \le \frac{c_1p^3}{48}$, then
$\eta_{ij}=0$ for all $ij \in E(G)$ implies that
$\eta_{ij} = 0$ for all $ij \in E(\Knn)$. 
For $ij \in E_b$, since $\delta_{ij} = \eta_{ij} = 0$, it follows that $\ell_{ij} = \ell_{ij}^{(0)}$.
Since $\delta_{ij} \|\toij\|_2 = \ell_{ij} - \ell_{ij}^{(0)}$ for $ij \in E_g$, we have
\[
	0
	= \sum_{ij \in E(G)} (\ell_{ij} - \ell_{ij}^{(0)})
	= \sum_{ij \in E_b} (\ell_{ij} - \ell_{ij}^{(0)}) + \sum_{ij \in E_g} (\ell_{ij} - \ell_{ij}^{(0)})
	= \sum_{ij \in E_g} (\ell_{ij} - \ell_{ij}^{(0)})
	= \sum_{ij \in E_g} \delta_{ij} \|\toij\|_2,
\] 
where the first equality is because $L(T, P) = L(\Tnot, \Pnot) = 1$.
By Condition 2, $\|\toij\|_2 \neq 0$ for all $i \neq j$. Therefore, if $\delta_{ij} \neq 0$ for some $ij \in E_g$, then there exists $ab, cd \in E_g$ 
such that $\delta_{ab} > 0$ and $\delta_{cd} < 0$. 
If $ab$ and $cd$ are vertex-disjoint, Lemma \ref{lem:rigidity4} and Conditions 2 and 4 force $\eta(ab,cd) > 0$, which contradicts the fact that
$\eta_{ij} = 0$ for all $ij \in E(\Knn)$.  If $ab$ and $cd$ are not vertex-disjoint, then, let $abc'd'$ be an arbitrary $C_4$ containing $ab$ and $cd$.  Then Lemma \ref{lem:rigidity4} implies the same result as above.
Therefore $\delta_{ij} = 0$ for all $ij \in E_g$, and hence
$\delta_{ij} = 0$ for all $ij \in E(G)$.  

Define $t_i = t_i^{(0)} + h_i$ for each $i \in \Vl$.  Define $p_j = \poj + h_j$ for $j \in \Vs$. Because $\eta_{ij} = \delta_{ij} = 0$ for all $ij \in E(G)$, we have $h_i = h_j$ 
for all $ij \in E(G)$. Since $G$ is connected by Condition 1, this implies $h_i = h_j$ for all 
$i \in \Vl, j \in \Vs$. Then by the constraint $\sum_{i \in \Vl} t_i + \sum_{j \in \Vs} p_j = 0$, 
we get $h_i = 0$ for all $i \in \Vl$ and $h_j = 0$ for all $j \in \Vs$. Therefore $T = \Tnot$ and $P = \Pnot$. \\

\noindent\textbf{Case 1}. $|\Eb| \neq 0$ and $\bar{\delta} \neq 0$ and $\sum_{ij\in E_{g}}|\delta_{ij}|<\frac{1}{8}\overline{\delta}|E_{g}|$. 
\medskip

Define $L_{b}=\{ij\in E_{b}:|\delta_{ij}|\ge\frac{1}{2}\overline{\delta}\}$.
Note that $\sum_{ij\in E_{b}\setminus L_{b}}|\delta_{ij}|<\frac{1}{2}\overline{\delta}|E_{b}|$
and therefore 
\begin{equation}
\sum_{ij\in L_{b}}|\delta_{ij}|=\sum_{ij\in E_{b}}|\delta_{ij}|-\sum_{ij\in E_{b}\setminus L_{b}}|\delta_{ij}|>\sum_{ij\in E_{b}}|\delta_{ij}|-\frac{1}{2}\overline{\delta}|E_{b}|=\frac{1}{2}\overline{\delta}|E_{b}|.\label{eq:large_delta-bipartite}
\end{equation}
Define $F_{g}=\{ij\in E_{g}:|\delta_{ij}|<\frac{1}{4}\overline{\delta}\}$.
Then by the condition of Case 1,
\[
\frac{1}{8}\overline{\delta}|E_{g}|>\sum_{ij\in E_{g}}|\delta_{ij}|\ge\sum_{ij\in E_{g}\setminus F_{g}}|\delta_{ij}|\ge\frac{1}{4}\overline{\delta}|E_{g}\setminus F_{g}|,
\]
and therefore $|E_{g}\setminus F_{g}|<\frac{1}{2}|E_{g}|$, or equivalently,
$|F_{g}|>\frac{1}{2}|E_{g}|$.

For each $ij\in L_{b}$, define $F_g(i,j) = \{k\ell \in F_g \mid k \neq i, \ell \neq j\}$. Note that by Condition 1, $|F_g(i,j)| > \frac{1}{2} |\Eg| - 2 p (\nl + \ns)$. For any $k\ell \in F_g(i,j)$, since $|\delta_{ij}| \ge \frac{1}{2} \overline{\delta}$ and $|\delta_{k\ell}| < \frac{1}{4}\overline{\delta}$, we have $|\delta_{ij} - \delta_{k\ell}| \ge \frac{1}{2}|\delta_{ij}|$. Thus
Lemma \ref{lem:rigidity4}
and Conditions 3 and 4 give $\eta(ij,k\ell)\ge \beta|\delta_{ij} - \delta_{k\ell}| \cdot\|\toij\|_2 \ge
\beta \cdot \frac{1}{2}|\delta_{ij}| \cdot \|\toij\|_2 \ge
\frac{\beta c_{0}\muinf}{2}|\delta_{ij}|$. 
Therefore by Condition 1, 
\begin{eqnarray*}
	\sum_{ij\in E_{b}}\sum_{\substack{k\ell\in E_{g}\\k\neq i, l \neq j}}\eta(ij,k\ell) 
	& \ge & \sum_{ij\in L_{b}}\sum_{k\ell\in F_{g}(i,j)}\frac{\beta c_{0}\muinf}{2}|\delta_{ij}|=\sum_{ij\in L_{b}}|F_{g}(i,j)|\cdot\frac{\beta c_{0}\muinf}{2}|\delta_{ij}|\\
	& >  & \sum_{ij\in L_{b}}\frac{\beta c_{0}\muinf}{2}\Bigl(\frac{1}{2}|E_{g}|  - 2p (\nl + \ns) \Bigr)|\delta_{ij}|
\end{eqnarray*}
Note that if $\eps < \frac{1}{4}p$, then $|\Eg| \geq \frac{\nl \ns p}{2} - |\Eb| \geq \frac{\nl \ns p}{4}.$
Further note that $\nl, \ns > 64$ implies that $2 p (\nl + \ns) < \frac{1}{16}\nl \ns p$.  Hence by \eqref{eq:large_delta-bipartite},
\begin{align*}
	\sum_{ij\in E_{b}}\sum_{\substack{k\ell\in E_{g}\\k\neq i, \ell \neq j}}\eta(ij,k\ell)
	& \ge\frac{\beta c_{0}\muinf}{32} \nl \ns \cdot \sum_{ij\in L_{b}} |\overline{\delta}_{ij}|
	\ge\frac{\beta c_{0}\muinf}{32} \nl \ns \cdot\frac{1}{2}\overline{\delta}|E_{b}|.
\end{align*}
For
each $ij\in E(\Knn)$, we would like to count how many times each
$\eta_{ij}$ appear on the left hand side. If $ij\in E_{b}$, then
there are at most $\nl \ns$ $C_{4}$s  containing
$ij$; hence $\eta_{ij}$ may appear at most $4 \nl \ns$
times. If $ij\notin E_{b}$, then $\eta_{ij}$ appears when there is
a $C_{4}$ containing $ij$ and some bad edge.  If the
bad edge is incident to $ij$, then there are at most
$2\eps \nl \ns$ such $C_{4}$s, and if the bad edge is
not incident to $ij$, then there are at most $|E_{b}|\le\eps \nl \ns$
such $C_{4}$s. Thus $\eta_{ij}$ may appear at most $4 \cdot 3 \eps \nl \ns = 12 \eps \nl \ns$ times. Therefore
\begin{eqnarray*}
	\sum_{ij\in E_{b}}\sum_{k\ell\in E_{g}}\eta(ij,k\ell) 
	& \le & \sum_{ij\in E_{b}}4 \nl \ns\cdot\eta_{ij}+\sum_{ij\in E(\Knn)}12 \varepsilon \nl \ns \cdot\eta_{ij}.
\end{eqnarray*}
By Lemma \ref{lem:badgood-bipartite}, if $\varepsilon < \frac{c_1p^3}{48}$, we have
\[
	\sum_{ij\in E_{b}}\sum_{k\ell\in E_{g}}\eta(ij,k\ell)
	\le\frac{48 \cdot 4\varepsilon}{c_{1}p^{3}}\nl \ns\sum_{ij\in E_{g}}\eta_{ij}+\sum_{ij\in E(\Knn)} 12 \eps \nl \ns\cdot\eta_{ij}
	\le\frac{204 \varepsilon}{c_{1}p^{3}}\nl \ns\sum_{ij\in E(\Knn)}\eta_{ij}.
\]
Hence
\[
	\frac{204\varepsilon}{c_{1}p^{3}}\nl \ns\sum_{ij\in E(\Knn)}\eta_{ij}\ge\frac{\beta c_{0}\muinf}{64} \nl \ns \cdot\overline{\delta}|E_{b}|.
\]

If $\varepsilon< \frac{\beta c_0 c_1^2 p^4}{384\cdot 204 \cdot 64}$, then by Condition 3,
 $\bar{\delta} \neq 0$, and $|\Eb| \neq 0$, the above implies 
\begin{align*}
	\sum_{ij\in E(\Knn)}\eta_{ij}
	\ge&\, \frac{\beta c_{0}c_{1}p^{3}}{204 \cdot 64 \eps}\muinf \cdot\overline{\delta}|E_{b}| \\
	>&\, \frac{384}{c_{1}p}\muinf\cdot\overline{\delta}|E_{b}|
	\ge \frac{384}{c_{1}p}\sum_{ij\in E_{b}}|\delta_{ij}|\|t_{ij}^{(0)}\|_2.
\end{align*}
Lemma \ref{lem:transference-bipartite} implies 
\[
	\sum_{ij\in E_{g}}\eta_{ij}
	\ge \frac{c_{1}p}{192}\sum_{ij\in E(\Knn)}\eta_{ij}
	> 2\sum_{ij\in E_{b}}|\delta_{ij}|\|t_{ij}^{(0)}\|_2.
\]
Therefore by (\ref{eq:badtwist-bipartite}),we have $\sum_{ij\in E_{g}}\eta_{ij}>\sum_{ij\in E_{b}}(|\delta_{ij}|\|t_{ij}^{(0)}\|_2+\eta_{ij})$ if $\eps \le \min\{\frac{c_1p^3}{48}, \frac{p}{4}, \frac{\beta c_0 c_1^2 p^4}{384 \cdot 204 \cdot 64} \}$ and $p \geq \sqrt{\frac{8}{n}}$.
By (\ref{eq:gain3-bipartite}), this shows $R(T,P)>R(\Tnot, \Pnot)$. This condition on $\eps$ is satisfied under the assumption $\eps \leq \frac{\beta c_0 c_1^2 p^4}{384 \cdot 204\cdot64}$.
\\

\noindent\textbf{Case 2}. $|\Eb| \neq 0$ and $\bar{\delta} \neq 0$ and $\sum_{ij\in E_{g}}|\delta_{ij}|\ge\frac{1}{8}\overline{\delta}|E_{g}|$. 
\medskip

Define $E_{+}=\{ij\in E_{g}\,:\,\delta_{ij}\ge0\}$ and $E_{-}=\{ij\in E_{g}\,:\,\delta_{ij}<0\}$.
Since $\ell_{ij}-\ell_{ij}^{(0)}=\delta_{ij}\|t_{ij}^{(0)}\|_2$
for $ij\in E_{g}$, we have 
\begin{eqnarray*}
0=\sum_{ij \in E(G)}(\ell_{ij}-\ell_{ij}^{(0)}) & = & \sum_{ij\in E_{b}}(\ell_{ij}-\ell_{ij}^{(0)})+\sum_{ij\in E_{g}}\delta_{ij}\|t_{ij}^{(0)}\|_2.
\end{eqnarray*}
where the first equality follows from $L(T,P) = L(\Tnot, \Pnot)$.  Therefore, 
\begin{eqnarray*}
	\left| \sum_{ij\in E_{g}}\delta_{ij}\|t_{ij}^{(0)}\|_2 \right|
	\le \left| \sum_{ij\in E_{b}}(\ell_{ij}-\ell_{ij}^{(0)}) \right|
	\le \sum_{ij\in E_{b}}(|\delta_{ij}|\|t_{ij}^{(0)}\|_2+\eta_{ij}) 
	\le 2\muinf\overline{\delta}|E_{b}|,
\end{eqnarray*}
where the last inequality follows from (\ref{eq:badtwist-bipartite}), Condition 3, and the definition of $\overline{\delta}$. On the other hand, the condition of Case 2 and Condition 3 implies 
$\sum_{ij\in E_{g}}|\delta_{ij}|\|t_{ij}^{(0)}\|_2 \ge \frac{1}{8}c_0 \muinf\overline{\delta}|E_g|$. 
Therefore
\[
	\sum_{ij \in E_-} (-\delta_{ij}) \|t_{ij}^{(0)}\|_2
	= \frac{1}{2}\left( -\sum_{ij\in E_{g}}\delta_{ij}\|t_{ij}^{(0)}\|_2 + \sum_{ij\in E_{g}}|\delta_{ij}|\|t_{ij}^{(0)}\|_2\right)
	\ge \frac{1}{2} \left(\frac{1}{8} c_0 \muinf\overline{\delta}|E_g| - 2\muinf\overline{\delta}|E_{b}|\right).
\]
If $\varepsilon \le \frac{1}{128}c_0p$, then 
since $|E_b| \le \varepsilon \nl \ns$ and $|E_g| \ge \frac{\nl \ns p}{2} - |E_b| \ge \frac{\nl \ns p}{4}$, 
we see that $\frac{1}{8} c_0 \muinf\overline{\delta}|E_g| - 2\muinf\overline{\delta}|E_{b}| \ge \frac{1}{16}c_0\muinf \bar{\delta}|E_g|$.
Therefore $\sum_{ij\in E_{-}}(-\delta_{ij})\|t_{ij}^{(0)}\|_2 \ge\frac{1}{32}c_{0}\muinf \overline{\delta}|E_{g}|$.
Similarly, $\sum_{ij\in E_{+}} \delta_{ij}\|t_{ij}^{(0)}\|_2 \ge \frac{1}{32}c_{0}\muinf \overline{\delta}|E_{g}|$.

If $|E_+| \ge \frac{1}{2}|E_g|$, then 
by Lemma \ref{lem:rigidity4} and Conditions 3 and 4, we have 
\begin{eqnarray*}
	\sum_{ij\in E_{-}}\sum_{\substack{k\ell\in E_{+}\\k\neq i, \ell \neq j}}\eta(ij,k\ell) 
	& \ge & \sum_{ij\in E_{-}}\sum_{\substack{k\ell\in E_{+}\\k\neq i, \ell \neq j}}\beta(-\delta_{ij})\|t_{ij}^{(0)}\|_2\\
	& \ge & \sum_{ij\in E_{-}}(-\delta_{ij})\|t_{ij}^{(0)}\|_2\cdot \beta(|E_{+}| - 2 p (\nl + \ns) ) \\
	& \ge & \frac{1}{32}c_{0}\muinf \overline{\delta}|E_{g}| \cdot \beta (|E_{+}| - 2 p (\nl + \ns) ) \\
	& \ge & \frac{\beta}{32} c_{0} \muinf\overline{\delta}|E_{g}| \Bigl(\frac{1}{2} | E_g| - 2 p (\nl + \ns) \Bigr).
\end{eqnarray*}
Note that if $\eps < \frac{1}{4}p$, then $|\Eg| \geq \frac{\nl \ns p}{2} - |\Eb| \geq \frac{\nl \ns p}{4}.$
Further note that $\nl, \ns > 64$ implies that $2 p (\nl + \ns) < \frac{1}{16}\nl \ns p$.  Hence,
\begin{eqnarray*}
	\sum_{ij\in E_{-}}\sum_{\substack{k\ell\in E_{+}\\k\neq i, l \neq j}}\eta(ij,k\ell) 
	&\ge & 
	\frac{1}{32}\beta c_0 \muinf \overline{\delta}\cdot \frac{\nl \ns p}{4} \cdot \frac{\nl \ns p}{16}
	\ge
	\frac{\beta c_0 \muinf \overline{\delta} \nl^2 \ns^2 p^2}{32 \cdot 64}.
\end{eqnarray*}
Similarly, if $|E_-| \ge \frac{1}{2}|E_g|$, then we can switch the order of summation and
consider $\sum_{ij \in E_+} \sum_{k\ell \in E_-} \eta(ij, k\ell)$ to obtain the 
same conclusion.

Since each edge is contained in at most $\nl \ns$
copies of $C_{4}$ and there are 4 edges
in a $C_{4}$, we have 
\[
	\sum_{ij\in E_{-}}\sum_{\substack{k\ell\in E_{+}\\k\neq i, \ell \neq j}}\eta(ij,k\ell)
	\le 4 \nl \ns \sum_{ij\in E(\Knn)}\eta_{ij}.
\]
If $\varepsilon < \frac{\beta c_0 c_1 p^3}{384\cdot4\cdot32\cdot64}$, then since $\bar{\delta} \neq 0$ and $|\Eb| \leq \eps \nl \ns$, we have
\[
	\sum_{ij\in E(\Knn)}\eta_{ij}
	\ge\frac{1}{4 \nl \ns}\cdot\frac{\beta c_{0} \muinf \overline{\delta}}{32\cdot 64} \nl^2 \ns^2 p^2
	\ge\frac{\beta c_{0} p^2}{4 \cdot 32 \cdot 64}\muinf \overline{\delta} \nl \ns
	>\frac{384}{c_{1}p}\muinf\overline{\delta}|E_{b}|.
\]
By Lemma \ref{lem:transference-bipartite}, if $\varepsilon < \frac{c_1p^3}{48}$, then this implies 
\[
\sum_{ij\in E_{g}}\eta_{ij}\ge\frac{c_{1}p}{192}\sum_{ij\in E(\Knn)}\eta_{ij} > 2\muinf\overline{\delta}|E_{b}|.
\]
Therefore from (\ref{eq:gain3-bipartite}), (\ref{eq:badtwist-bipartite}), and Condition 3, 
if $\varepsilon \le \min\{\frac{c_0 p}{128}, \frac{c_1p^3}{96}, \frac{p}{4}, \frac{\beta c_0 c_1 p^3}{384\cdot4\cdot32\cdot64} \}$ and $p \geq \sqrt{\frac{8}{n}}$, then
\begin{eqnarray*}
R(T, P) & \ge & R(\Tnot, \Pnot)+\sum_{ij\in E_{g}}\eta_{ij}-\sum_{ij\in E_{b}}(|\delta_{ij}|\|t_{ij}^{(0)}\|_2+\eta_{ij})\\
 & > & R(\Tnot, \Pnot)+2\muinf\overline{\delta}|E_{b}|-\sum_{ij\in E_{b}}2|\delta_{ij}|\|t_{ij}^{(0)}\|_2\ge R(\Tnot, \Pnot).
\end{eqnarray*}
This condition on $\eps$ is satisfied under the assumption $\eps \leq \frac{\beta c_0 c_1^2 p^4}{384\cdot204\cdot64}$.
\end{proof}

\subsection{Properties of Gaussians} \label{sec-properties-of-gaussians}

In this section, we prove that i.i.d. Gaussians satisfy properties needed to establish Conditions 3--5 in Theorem \ref{thm:mainthm}. We begin by recording some useful facts regarding concentration of random Gaussian vectors:

\begin{lemma}
\label{lem:conc1}
Let $x,y$ be i.i.d. $\mathcal{N}(0,I_{d\times d})$, and $\epsilon \leq 1$, then
\[
\P\left ( d(1-\epsilon) \leq \|x\|_2^2 \leq d(1+\epsilon) \right) \geq 1- e^{-c \epsilon^2 d}
\]
and
\[
\P \left( |\langle x, y \rangle| \geq d\epsilon \right) \leq e^{-c\epsilon^2 d}
\]
where $c>0$ is an absolute constant.
\end{lemma}
\begin{proof}
Both statements follow from Corollary 5.17 in \cite{Vershynin}, concerning concentration of sub-exponential random variables.
\end{proof}

\begin{lemma}
\label{lem:conc2}
Corollary 5.35 in \cite{Vershynin}. Let $A$ be an $n \times d$ matrix with iid $\mathcal{N}(0,1)$ entries. Then for any $t\geq 0$,
\[
\P\left( \sigma_{\text{max}}(A) \geq \sqrt{n} + \sqrt{d} + t \right) \leq 2e^{-\frac{t^2}{2}}
\]
where $\sigma_{\max}(A)$ is the largest singular value of $A$.
\end{lemma}

\begin{lemma} \label{lem-beta-first}
Let $\ti, \tj, \tk, \tl \sim \mathcal{N}(0, I_{d \times d})$ be independent.  There is a universal constant $c$ such that with probability at least $1 - 15 e^{-cd}$,
$$
\frac{\|\proj_{\Span(\tk - \tj, \ti - \tl)^\perp} (\ti - \tj) \|}{\| \ti - \tj\|} \geq \frac{1}{4}.
$$
\end{lemma}
\begin{proof}
Let $c$ be the constant from Lemma \ref{lem:conc1}.
Let $x = \ti - \tl, y = \tk -\tj , z = \ti - \tj$.  Observe
\begin{align*}
\proj_{\Span(x,y)^\perp} \zhat  
= \zhat - \langle \zhat, \xhat \rangle \xhat - \langle \zhat, \yxphat \rangle \yxphat 
=&\, \zhat - \langle \zhat, \xhat \rangle \xhat - \langle \zhat, \yhat \rangle \yhat + (\langle \zhat, \yhat \rangle \yhat - \langle \zhat, \yxphat \rangle \yxphat) \\
  =&\, \zhat - \langle \zhat, \xhat \rangle \xhat - \langle \zhat, \yhat \rangle \yhat +(\yhat \yhat^t - \yxphat \yxphat^t ) \zhat,
\end{align*}
where $\yxphat = \frac{y - \langle y, \xhat \rangle \xhat}{\|y - \langle y, \xhat \rangle \xhat \|}$, which is well defined with probability $1$.  By the triangle inequality,
$$
\| \proj_{\Span(x,y)^\perp} \zhat \| \geq \sqrt{1 - | \langle \zhat, \xhat \rangle |^2} - | \langle \zhat, \yhat \rangle |  - \| \yhat \yhat^t - \yxphat \yxphat^t \|_{\text{op}}
$$
For arbitrary unit vectors $\ahat, \bhat \in \R^d$,  $\| \ahat \ahat^t - \bhat \bhat^t\|_\text{op} =  |\sin \theta|$, where $\theta$ is the angle between $\ahat$ and $\bhat$.  This fact can be verified by direct computation after taking $\ahat = e_1$ and $\bhat = \cos \theta \ e_1 + \sin \theta \ e_2$ without loss of generality.  Hence, $\| \yhat \yhat^t - \yxphat \yxphat^t\|_\text{op} =  | \sin \theta | =  | \cos \alpha |$, where $\theta$ is the angle between $\yhat$ and $\yxphat$, and $\alpha$ is the angle between $\yhat$ and $\xhat$.  Thus $\| \yhat \yhat^t - \yxphat \yxphat^t\|_\text{op} = | \langle \yhat, \xhat \rangle|$. 
So, 
$$
\| \proj_{\Span(x,y)^\perp} \zhat \| \geq \sqrt{1 - | \langle \zhat, \xhat \rangle |^2} - | \langle \zhat, \yhat \rangle | - | \langle \yhat, \xhat \rangle |
$$
Now, note that 
$$
| \langle \zhat, \xhat \rangle |^2 = \frac{\langle \ti - \tj, \ti - \tl \rangle^2}{\| \ti - \tj \|^2 \| \ti - \tl \|^2} = \frac{(\|\ti\|^2 - \langle \ti, \tl \rangle - \langle \tj, \ti \rangle + \langle \tj, \tl \rangle)^2}{\| \ti - \tj \|^2 \| \ti - \tl \|^2}
$$
By Lemma \ref{lem:conc1} with $\eps = 0.01$,
$$
| \langle \zhat, \xhat  \rangle |^2 \leq \frac{(d ( 1+ \eps) + 3d \eps )^2}{4 d^2 (1-\eps)^2} \leq 0.3
$$
with probability at least $1 - 6 e^{-c d}$ for some universal constant $c$.   Similarly, $| \langle \zhat, \yhat \rangle |^2 \leq 0.3$ with the same probability.
Since $\yhat$ and $\xhat$ are independent, by Lemma \ref{lem:conc1} with $\eps = 0.01$, $| \langle \yhat, \xhat \rangle | \leq \frac{\eps d}{d (1-\eps)} \leq 2 \eps$ with probability at least $1 - 3 e^{-c d}$.
Thus, we observe 
$$
\| \proj_{\Span(x,y)^\perp} \zhat \| \geq \sqrt{1 - 0.3} - \sqrt{0.3} - 0.02 \geq \frac{1}{4}
$$
with probability at least $1 - 15 e^{-c d}$. 
\end{proof}

\begin{lemma} \label{lem-beta-second}
Let $\ti, \tj, \tk, \tl \sim \mathcal{N}(0, I_{d \times d})$ be independent for $d \geq 3$.  There is a universal constant $c$ such that with probability at least $1 - 7 e^{-cd}$,
$$
\frac{\|\proj_{\Span(\ti - \tl, \tk - \tl)^\perp} (\ti - \tj) \|}{\|\ti - \tj \|_2} \geq \frac{1}{4}.
$$
\end{lemma}
\begin{proof}
Let $c$ be the constant from Lemma \ref{lem:conc1}.
Let $u = \frac{\ti - \tl}{\sqrt{2}}, v = \frac{\tl + \ti}{\sqrt{2}}, w = \tj, x = \tk$.  Each of these variables are i.i.d. $\mathcal{N}(0, 1)$.
Note that 
$$
\proj_{\Span(\ti - \tl, \tk - \tl)^\perp} (\ti - \tj) = -\proj_{\Span(u, x - \frac{v}{\sqrt{2}})^\perp} \Bigl(w - \frac{v}{\sqrt{2}} \Bigr).
$$
Without loss of generality, rotate coordinates so that $u$ is in the direction of $e_1$.  Thus, it suffices to bound
$
\bigl \| \proj_{(\xtilde - \frac{\vtilde}{\sqrt{2}})^\perp} ( \wtilde - \frac{\vtilde}{\sqrt{2}} ) \bigr \|_2
$
where $\vtilde, \wtilde, \xtilde \sim \mathcal{N}(0, I_{d-1 \times d-1})$.  Note that $\xtilde - \frac{\vtilde}{\sqrt{2}}$ and $\wtilde - \frac{\vtilde}{\sqrt{2}}$ both follow the distribution $\mathcal{N}(0,\frac{3}{2} I_{d-1 \times d-1})$.
Note that 
\begin{align*}
\Bigl \| \proj_{(\xtilde - \frac{\vtilde}{\sqrt{2}})^\perp} \Bigl( \wtilde - \frac{\vtilde}{\sqrt{2}} \Bigr) \Bigr \|^2 
&= \Bigl \|\wtilde - \frac{\vtilde}{\sqrt{2}} \Bigr \|^2 - \frac{\langle \wtilde - \frac{\vtilde}{\sqrt{2}}, \xtilde - \frac{\vtilde}{\sqrt{2}} \rangle^2}{\| \xtilde - \frac{\vtilde}{\sqrt{2}}\|^2} \\
&= \Bigl \|\wtilde - \frac{\vtilde}{\sqrt{2}} \Bigr \|^2 - \frac{\bigl (\langle \wtilde, \xtilde \rangle - \frac{1}{\sqrt{2}} \langle \wtilde, \vtilde \rangle - \frac{1}{\sqrt{2}} \langle \vtilde, \xtilde \rangle + \frac{\| \vtilde\|^2}{2} \bigr)^2}{\| \xtilde - \frac{\vtilde}{\sqrt{2}}\|^2}.
\end{align*}

Hence Lemma \ref{lem:conc1} with $\eps = 0.01$ shows that  with probability at least $1 - 6 e^{-c d}$, the above is at least
\begin{align*}
\frac{3}{2} (d-1) (1-\eps) - \frac{\Bigl(\frac{1}{2} (d-1) (1+\eps) - 3 \eps (d-1)\Bigr)^2}{\frac{3}{2} (d-1) (1-\eps)} \geq (d-1) \geq \frac{2}{3} d,
\end{align*}
Thus, we have that 
$$
\|\proj_{\Span(\tj - \tl, \tk - \tl)^\perp} (\ti - \tj) \|^2  \geq \frac{2}{3} d
$$
with probability at least $1 - 6 e^{-c d}$.  To conclude the proof, note that Lemma \ref{lem:conc1} with $\eps = 0.01$ implies that $\|\ti - \tj\|^2 \geq 2 d ( 1+\eps)$ with probability at least $1 - e^{-c d}$.
\end{proof}

We can now establish Conditions 3--4 of Theorem~\ref{thm:mainthm} with high probability.
\begin{lemma}\label{lem-conds34}
Let $\ti, \pj \sim \mathcal{N}(0, I_{d \times d})$ for $i \in \Vl, j \in \Vs$ be independent.   Condition 3 of Theorem~\ref{thm:mainthm} holds with $c_0 = \frac{9}{10}$ with probability at least $1 - 2 \nl \ns e^{-cd}$ for a universal constant $c$.  Condition 4 of Theorem~\ref{thm:mainthm} holds with $\beta = \frac{1}{4}$ with probability at least $ 1 - 22 \nl^2 \ns^2 e^{-c d}$. 
\end{lemma}
\begin{proof}
Condition 3 follows from applying Lemma \ref{lem:conc1} with $\eps=0.01$ and a union bound to $\|\ti - \pj\|_2$ for all $\nl \ns$ pairs $(i,j) \in \Vl \times \Vs$.  Condition 4 follows from applying Lemmas \ref{lem-beta-first} and \ref{lem-beta-second} and a union bound over the at most $\nl^2 \ns^2$ choices of $i, k \in \Vl$ and $j, \ell \in \Vs$. 
\end{proof}

\subsection{Gaussians are well-distributed} \label{sec-gaussians-well-distributed}

In this section we prove that Condition 5 of Theorem~\ref{thm:mainthm} holds with high probability.

\begin{lemma} \label{lem-c4-well-distributed}
There exist constants $d_0$ and $K_0$ such that the following holds.  Let $G = (\Vl \cup \Vs, E)$ be a bipartite-$p$-typical graph.  Let $\ti, \pj \sim \mathcal{N}(0, I_{d \times d})$ for $i \in \Vl, j \in \Vs$ be independent from $G$ and each other.  Let $T = \{\ti\}_{i \in \Vl}$ and $P = \{\pj\}_{j \in \Vs}$.  If $d \geq d_0$ and $\nl, \ns \geq \max(K_0, 160 d)$, then $(T, P)$ is $\frac{1}{20}$-well-distributed along $G$ with probability at least $1 - O( \nl^2 \ns^2 e^{-cd})$ for  universal constants $c, K_0$.
\end{lemma}

We start by proving an intermediate lemma asserting the well-distributedness
of pairs of random Gaussian vectors $\{(t_i, p_i)\}_{i \in [k]}$ with respect to a fixed pair of random Gaussian vectors $(x,y)$.

\begin{lemma} \label{lem:wd_over_matching}
There exist positive constants $d_0,\Kotilde$ such that the following holds. Let $x,y, \ti, p_i \sim \mathcal{N}(0, I_{d \times d})$ be independent, where $i \in [k]$. Then the set $\{(t_i, p_i)\}_{i \in [k]}$
is $\frac{1}{10}$-well-distributed with respect to $(x,y)$ with probability $1 - 6 k e^{-cd}$ if $k \geq \max(\Kotilde, 10d)$ and $d \geq d_0$.
\end{lemma}

The proof of this lemma appears at the end of this section.
We will deduce Lemma~\ref{lem-c4-well-distributed} from Lemma~\ref{lem:wd_over_matching}
by partitioning the edge set of $G$ into sets of vertex-disjoint edges.
A {\em matching} is a set of vertex-disjoint edges.
A {\em perfect matching} of a graph is a matching that intersects all vertices.
The following is a well-known lemma in Graph theory.

\begin{lemma} \label{lem:decompose}
Let $G = (V, E)$ be a bipartite graph with vertex partition $V = V_1 \cup V_2$, and
let $\Delta$ be the maximum degree of $G$. 
There exists an edge-partition $E = E_1 \cup \cdots \cup E_\Delta$ such that $E_a$ forms a matching
for each $a \in [\Delta]$.
\end{lemma}
\begin{proof}
By adding vertices and edges to $G$ if necessary, we can obtain a $\Delta$-regular 
bipartite multi-graph $G'$.
By Hall's theorem, every non-empty
regular multi-graph contains a perfect matching (see \cite[Corollary 2.1.3]{Diestel}). 
Let $F_1$ be an arbitrary perfect matching of $G'$. Remove $F_1$ from the edge set of $G'$, 
and note that the remaining graph is still regular. Thus we can repeat the process to 
obtain a partition $E(G') = F_1 \cup \cdots \cup F_{\Delta}$ of the edge set of $G'$ into perfect 
matchings. 
The sets $E_a = F_a \cap E(G)$ for $a \in [\Delta]$ satisfy the claimed condition.
\end{proof}

The proof of Lemma~\ref{lem-c4-well-distributed} follows from the two lemmas above.

\begin{proof}[Proof of Lemma~\ref{lem-c4-well-distributed}]
Recall the notation that $N = \max\{|\Vl|, |\Vs|\}$ and $n = \min\{|\Vl|, |\Vs|\}$.  Since $G$ is a bipartite-$p$-typical graph, the maximum degree $\Delta$ of $G$ is at most
$2N p$. By Lemma~\ref{lem:decompose}, there exists an edge-partition 
$E = E_1 \cup \cdots \cup E_\Delta$ such that each $E_a$ for $a=1,2,\ldots, \Delta$ forms
a matching.

Fix a pair of indices $(i_0,j_0)$ for $i_0 \in \Vl$ and $j_0 \in \Vs$.
Let $E' \subseteq E$ be the subset of edges that do not intersect $i_{0}$ or $j_{0}$, and
for each $a \in [\Delta]$, let $E_a' \subseteq E_a$ be the subset of edges that
do not intersect $i_{0}$ or $j_{0}$.
Let $A \subseteq [\Delta]$ be the set of indices $a$ for which $|E_a'| \ge \max(\tilde{K}_0, 10 d)$.
For each $a \in A$, by Lemma~\ref{lem:wd_over_matching}, we see that 
with probability at least $1 - O(|E_a'| e^{-cd})$,
\[
	\sum_{ij \in E_a'} \| \proj_{\Span\{p_{j_0}-t_{i},t_{i}-p_{j}, p_j-t_{i_0}\}^{\perp}} (h) \|_2
	\ge \frac{1}{10} |E_a'| \| h \|_2
\]
holds for all $h \in \mathbb{R}^d$.
Therefore by the union bound, with probability at least $1 - O(\sum_{a \in A} |E_a'| e^{-cd}) \geq 1 - O( |E'| e^{-cd}) \geq 1 - O( nN e^{-cd})$, the
above holds simultaneously for all $a \in [\Delta]$. Conditioned on this event, for all $h \in \mathbb{R}^d$,
\begin{align*}
	&\, \sum_{ij \in E'} \| \proj_{\Span\{p_{j_0}-t_{i},t_{i}-p_{j}, p_j-t_{i_0}\}^{\perp}} (h) \|_2 \\
	\ge &\,
	\sum_{a \in A} \sum_{ij \in E_a'} \| \proj_{\Span\{p_{j_0}-t_{i},t_{i}-p_{j}, p_j-t_{i_0}\}^{\perp}} (h) \|_2
	\ge \sum_{a \in A} \frac{1}{10} |E_a'| \| h \|_2.
\end{align*}
Since $|E'| = \sum_{a =1}^{\Delta} |E_a'|$, we see that
\begin{align} \label{eq:combine_rotation}
	\sum_{ij \in E'} \| \proj_{\Span\{p_{j_0}-t_{i},t_{i}-p_{j}, p_j-t_{i_0}\}^{\perp}} (h) \|_2
	\ge \frac{1}{10} \left( |E'| - \sum_{a \notin A} |E_a'| \right) \| h \|_2.
\end{align}
Since $G$ is bipartite-$p$-typical, we have $|E'| \ge \frac{1}{2}nNp - 2(N+n)p \ge \frac{1}{4}nNp$ if $n > 16$, and by the  definition of $A$, we have
$\sum_{a \notin A} |E_a'| \le  \max(\tilde{K}_0, 10 d) \cdot \Delta \le \frac{1}{8}nNp$ if $n \geq 16 \cdot \max(\tilde{K}_0, 10 d)$.
Hence the right-hand-side of \eqref{eq:combine_rotation} is at least
$\frac{1}{20}|E'| \| h \|_2$ for all $h \in \mathbb{R}^d$. This shows that the set $\{(\ti,\pj)\}_{i \neq i_0, j \neq j_0}$
is $\frac{1}{20}$-well-distributed with respect to $(t_{i_0}, p_{j_0})$ with probability 
at least $1 - O(n N e^{-cd})$. By taking the union bound over all 
choices of pairs $(i_0, j_0) \in \Vl \times \Vs$, we can conclude that $(T,P)$ is
$\frac{1}{20}$-well-distributed along $G$ with probability at least 
$1 - O(n^2 N^2 e^{-cd})$.
\end{proof}

We now prove Lemma~\ref{lem:wd_over_matching}.

\begin{proof}[Proof of Lemma~\ref{lem:wd_over_matching}]

Throughout the proof, the positive constant $c$ may change from line to line, but is always bounded below by the positive constant of the lemma statement.

For each $i$, let $
W_i = \Span(t_i - y, p_i -x, t_i - p_i) =\Span( x-y, p_i + t_i - (x+y), t_i-p_i).
$
Thus $P_{W_i^\perp} \circ P_{(x-y)^\perp} = P_{W_i^\perp}$. Therefore, it is enough to show that for all $h \perp x-y$, with high probability
\[
\sum_{i=1}^n \| P_{W_i^\perp}( h) \|_2 \geq \frac{1}{10}n \|h\|_2.
\]

Letting $V_i = \Span(x-y, p_i + t_i - x -y)$, we have
\[
W_i = \Span( x-y, p_i + t_i - x-y, t_i-p_i) = \Span \Bigl( x-y, p_i + t_i - x-y, P_{V_i^\perp}(t_i-p_i) \Bigr).
\]
Now, for any $h \perp (x-y)$, 
\begin{align*}
\sum_{i=1}^n \| P_{W_i^\perp}(h)\|_2  
& \geq \left\| \sum_{i=1}^n P_{W_i^\perp}( h) \right\|_2 \\
& = \left\| \sum_{i=1}^n \left(P_{V_i^\perp}( h) - P_{P_{V_i^\perp}(t_i-p_i)}(h) \right) \right\|_2 \\
& \geq \left\| \sum_{i=1}^n P_{V_i^\perp}( h)\right\|_2  - \left\| \sum_{i=1}^n P_{P_{V_i^\perp}(t_i-p_i)}(h)  \right\|_2 \\
& \geq \left\| \sum_{i=1}^n P_{V_i^\perp}( h)\right\|_2  - \sum_{i=1}^n \left\| P_{P_{V_i^\perp}(t_i-p_i)}(h) - P_{(t_i-p_i)}(h) \right\|_2 - \left\|\sum_{i=1}^n P_{(t_i-p_i)}(h)\right\|_2.
\end{align*}

Since $ \|P_v(h) - P_w (h) \|_2 \leq \| \vhat \vhat^t - \what \what^t\|_\text{op} \|h\|_2  \leq  \| \hat{v} - \hat{w} \|_2 \|h\|_2 $ holds for all vectors $v,w,h \in \mathbb{R}^d$, the above is at least
\begin{align} \label{eq:temp_midstep}
& \left\| \sum_{i=1}^n P_{V_i^\perp}( h)\right\|_2 - \|h\|_2 \sum_{i=1}^n \left \| \frac{P_{V_i^\perp}(t_i-p_i)}{\|P_{V_i^\perp}(t_i-p_i)\|_2} - \frac{(t_i-p_i)}{\|t_i - p_i\|_2}  \right \|_2 - \left\|\sum_{i=1}^n P_{(t_i-p_i)}(h)\right\|_2.
\end{align}
Note that when $v = P(w)$ for some orthogonal projection operator $P$, we have
\[
\| \hat{v} - \hat{w}\|_2^2 = 2\left(1-\langle \hat{v},\hat{w} \rangle\right) = 2\left(1 - \frac{\langle  P(w), w \rangle}{\|P(w)\|_2 \|w\|_2}\right) = 2\left(1- \frac{\|P(w)\|_2}{\|w\|_2}\right).
\]
Thus,
\[
\left \| \frac{P_{V_i^\perp}(t_i-p_i)}{\|P_{V_i^\perp}(t_i-p_i)\|_2} - \frac{(t_i-p_i)}{\|t_i - p_i\|_2}  \right \|_2 = \sqrt{2}\sqrt{1 - \frac{\| P_{V_i^\perp} (t_i-p_i) \|_2}{\|t_i-p_i\|_2}}.
\]
Hence \eqref{eq:temp_midstep} is at least
\begin{align*}
& \left\| \sum_{i=1}^n P_{V_i^\perp}( h)\right\|_2 -  \|h\|_2 \sum_{i=1}^n \left \| \frac{P_{V_i^\perp}(t_i-p_i)}{\|P_{V_i^\perp}(t_i-p_i)\|_2} - \frac{(t_i-p_i)}{\|t_i - p_i\|_2}  \right \|_2 - \frac{1}{\min_i(\|t_i - p_i\|_2^2)}\left \|\sum_{i=1}^n (t_i-p_i)(t_i-p_i)^*\right\|_{\text{op}} \|h\|_2\\
&= \left\| \sum_{i=1}^n P_{V_i^\perp}( h)\right\|_2 -  \|h\|_2 \sum_{i=1}^n \sqrt{2}\sqrt{1 - \frac{\| P_{V_i^\perp} (t_i-p_i) \|_2}{\|t_i-p_i\|_2}} - \frac{1}{\min_i(\|t_i - p_i\|_2^2)}\left \|\sum_{i=1}^n (t_i-p_i)(t_i-p_i)^*\right\|_{\text{op}} \|h\|_2.
\end{align*}
We will now expand the first term, $\| \sum_{i=1}^n P_{V_i^\perp}( h)\|_2$. Let $u = x-y$ and $z_i = x+y - (p_i + t_i)$. We have
\begin{align*}
P_{V_i^\perp}( h) &= P_{S(u,z_i)^\perp}(h)\\
&= P_{S(u,P_{u^\perp} (z_i))^\perp} (h) \\
& = h - P_{u}(h) - P_{P_{u^\perp} (z_i)}(h) \\
& = h - P_{P_{u^\perp} (z_i)}(h) + P_{z_i} (h) - P_{z_i} (h)\\
& =  P_{z_i^\perp} (h) - P_{P_{u^\perp} (z_i)}(h) + P_{z_i} (h),
\end{align*}
where we used $h \perp u$ in the fourth inequality. 
Thus,
\begin{align*}
\| \sum_{i=1}^n P_{V_i^\perp}( h)\|_2 &= \| \sum_{i=1}^n \left(P_{z_i^\perp}(h)  - P_{P_{u^\perp} (z_i)}(h) + P_{z_i} (h) \right) \|_2 \\
&\geq \| \sum_{i=1}^n P_{z_i^\perp}(h)\|_2 - \sum_{i=1}^n \| P_{P_{u^\perp} (z_i)}(h) - P_{z_i} (h) \|_2 \\
& \geq \| \sum_{i=1}^n P_{z_i^\perp}(h)\|_2 - \|h\|_2 \sum_{i=1}^n \left \| \frac{P_{u^\perp} (z_i)}{\|P_{u^\perp} (z_i)\|_2} - \frac{z_i}{\|z_i\|_2} \right \|_2 \\
& = \| \sum_{i=1}^n P_{z_i^\perp}(h)\|_2 - \|h\|_2 \sum_{i=1}^n \sqrt{2} \sqrt{1- \frac{\|P_{u^\perp} (z_i)\|_2}{\|z_i\|_2}}
\end{align*}

Letting $X_i = \frac{\| P_{V_i^\perp}(t_i-p_i) \|_2}{\|t_i-p_i\|_2}$, $Y_i = \frac{\| P_{(x-y)^\perp} (x+y - t_i-p_i) \|_2}{\|x+y - t_i-p_i\|_2}$, and $Z_i = \left \|\sum_{i=1}^n (t_i-p_i)(t_i-p_i)^*\right\|_{\text{op}}$, we have shown that for any $h \perp x-y$,
\begin{align}
\sum_{i=1}^n \| P_{W_i^\perp}(h)\|_2  \notag
\geq \sum_{i=1}^n &\left \| P_{(x+y-t_i - p_i)^\perp}(h)\right \|_2 \\&-  \|h\|_2 \sum_{i=1}^n \sqrt{2}\left[ \sqrt{1 - X_i} + \sqrt{1 - Y_i} \right]  - \frac{1}{\min_i(\|t_i - p_i\|_2^2)}Z_i\|h\|_2 \label{pwi-perp}
\end{align}
We will separately bound the first term and last two terms with high probability.

We now show that the first term of \eqref{pwi-perp} is bounded below by $0.3 n \|h\|_2$ with high probability.

Because $t_i + p_i =^d \sqrt{2}t_i$, it suffices to show that with high probability
\[
\left \| \sum_{i=1}^n P_{(x+y-\sqrt{2}t_i)^\perp}(h) \right\|_2 \geq   0.3 n\|h\|_2.
\]
Let $v = x+y$ and $w_i = -\sqrt{2}t_i$. Note that

\begin{align*}
\left \|  \sum_{i=1}^n P_{(v+w_i)^\perp}(h)\right\|_2 
&= \left \| \sum_{i=1}^n \left(h - \frac{1}{\|v+w_i\|_2^2}\langle h, v+ w_i \rangle (v+ w_i) \right)\right \|_2\\
&\geq n\|h\|_2 - \left \| \sum_{i=1}^n \frac{1}{\|v+w_i\|_2^2} (v+ w_i)(v+ w_i)^* h\right \|_2  \\
&\geq n\|h\|_2 - \left \| \sum_{i=1}^n \frac{1}{\|v+w_i\|_2^2} (v+ w_i)(v+ w_i)^* \right\|_{\text{op}} \|h\|_2\\
& \geq \|h\|_2 \left[n - \frac{1}{\min_{i}\|v+w_i\|_2^2} \left \| \sum_{i=1}^n (v+ w_i)(v+ w_i)^* \right\|_{\text{op}} \right],
\end{align*}
where in the last inequality we used 
\[
\sum_{i=1}^n \frac{1}{\|v+w_i\|_2^2} (v+ w_i)(v+ w_i)^* \preceq \frac{1}{\min_{i}\|v+w_i\|_2^2}  \sum_{i=1}^n (v+ w_i)(v+ w_i)^*.
\]
Now, let $A = \sum_{i=1}^n e_i w_i^* \in \mathbb{R}^{n\times d}$. We have  
\begin{align*}
\left \| \sum_{i=1}^n (v+ w_i)(v+ w_i)^* \right\|_{\text{op}} &= \left \| \sum_{i=1}^n (vv^* + vw_i^* + w_i v^* + w_i w_i^*) \right\|_{\text{op}}\\
& \leq n\|vv^*\|_{\text{op}} + \left\| v \left( \sum_{i=1}^n w_i \right)^* +  \left( \sum_{i=1}^n w_i \right)v^* \right\|_{\text{op}} + \left \| \sum_{i=1}^n w_i w_i^* \right\|_{\text{op}}\\
& \leq n \|v\|_2^2 + 2\|v\|_2\left \| \sum_{i=1}^n w_i \right\|_2 + \left \| \sum_{i=1}^n w_i w_i^* \right\|_{\text{op}}\\
& = n \|v\|_2^2 + 2\|v\|_2\left \| \sum_{i=1}^n w_i \right\|_2 + \sigma_{\max}(A)^2.
\end{align*}
Thus,
\[
\sum_{i=1}^n \left \| P_{(v+w_i)^\perp}(h)\right \|_2 \geq \|h\|_2\left[n - \frac{n \|v\|_2^2 + 2\|v\|_2\left \| \sum_{i=1}^n w_i \right\|_2 + \sigma_{\max}(A)^2}{\min_{i}\|v+w_i\|_2^2} \right].
\]

Now, consider the event
\[
E_1 = \left\{ \min_{i} \|v+w_i\|_2^2 \geq 4d\beta_1, \quad \|v\|_2^2 \leq 2d\beta_2, \quad \left \| \sum_{i=1}^n w_i \right\|_2^2 \leq 2nd\beta_3, \quad \sigma_{\max}(A)^2 \leq 2n\beta_4  \right\}
\]
On $E_1$ we have,
\begin{align*}
\sum_{i=1}^n \left \| P_{(v+w_i)^\perp}(h)\right \|_2 &\geq \|h\|_2 \left[ n - \frac{1}{4d\beta_1}\left( 2nd\beta_2 + 2\sqrt{2d\beta_2}\sqrt{2}\sqrt{nd}\sqrt{\beta_3} + 2n\beta_4  \right) \right]\\
&= \|h\|_2 \left[n - \frac{1}{2}n \frac{\beta_2}{\beta_1} - \frac{4d\sqrt{n} \sqrt{\beta_2\beta_3}}{4d\beta_1}-\frac{\beta_4 }{2d\beta_1}n \right]\\
& = \|h\|_2 \left[n\left(1 - \frac{1}{2} \frac{\beta_2}{\beta_1} - \frac{\beta_4}{2d\beta_1}- \frac{1}{\sqrt{n}}\frac{ \sqrt{\beta_2\beta_3}}{\beta_1} \right)\right]
\end{align*}
Now, let $\beta_1 = 1-\frac{1}{100}, \quad \beta_2 = \beta_3 = 1 + \frac{1}{100}, \quad \beta_4 = \frac{1}{5}d \beta_1$.  This  gives
\[
\frac{1}{2} \frac{\beta_2}{\beta_1} = 1/2 +1/99, \quad \frac{\beta_4}{2d\beta_1} = 1/10,  \quad \frac{1}{\sqrt{n}}\frac{  \sqrt{\beta_2\beta_3}}{\beta_1} < \frac{2}{\sqrt{n}}.
\]
Assuming $n \geq 550$, we see that on $E_1$,
\[\sum_{i=1}^n \left \| P_{(x+y-t_i - p_i)^\perp}(h)\right \|_2 \geq 0.3 n \|h\|_2. \]

Now, we bound $\P(E_1)$. 
Note that $ \frac{1}{4}\|v+w_i\|_2^2 =^d \frac{1}{2}\|v\|_2^2 =^d \frac{1}{2n} \left \| \sum_{i=1}^n w_i \right\|_2^2 =^d \chi^2(d)$ and $\frac{1}{\sqrt{2}}A$ is a random $n\times d$ matrix with i.i.d. $\mathcal{N}(0,1)$ entries. Thus, by applying Lemma \ref{lem:conc1}, we have 
\[
\P \Bigl( 4d(1-\epsilon) \leq \|v+w_i\|_2^2 \leq 4d(1+\epsilon) \Bigr) \geq 1 - e^{-c \epsilon^2 d}
\]

\[
\P\left( 2d(1-\epsilon) \leq \|v\|_2^2 \leq 2d(1+\epsilon) \right) \geq 1 - e^{-c \epsilon^2 d}
\]

\[
\P\left( 2nd(1-\epsilon) \leq \left \| \sum_{i=1}^n w_i \right\|_2^2 \leq 2nd(1+\epsilon) \right) \geq 1 - e^{-c \epsilon^2 d},
\]
where $c>0$ is a universal constant. Also by taking $t =2  \sqrt{d}$ in Lemma \ref{lem:conc2} we get
\[
\P \left( \sigma_{\text{max}}\Bigl(\frac{1}{\sqrt{2}} A\Bigr) \geq \sqrt{n} +3\sqrt{d} \right) \leq 2e^{-2d}
\]
We have
\[
\P  \left(\sigma_{\max} (\frac{1}{\sqrt{2}} A) \geq \sqrt{n \beta_4}\right) \leq \P\left( \sigma_{\max} (\frac{1}{\sqrt{2}} A) \geq \sqrt{n} + 3\sqrt{d} \right) \leq 2e^{-2d}
\]
whenever $\sqrt{n} + 3\sqrt{d} \leq \sqrt{n \beta_4}$, or equivalently $(\sqrt{\frac{\beta_1}{5}}\sqrt{d} -1) \sqrt{n} \geq 3 \sqrt{d}$,  
which holds when $n \geq 550$ and $d \geq 10$. Thus for $n\geq 550$, we have
\[
\mathbb{P}(E_1) \geq 1- 2n e^{-cd} .
\]

We now show that the second term of \eqref{pwi-perp} is bounded above by $0.2 n \|h\|_2$ with high probability.
Define the event
\[
E_2 = \left\{ X_i \geq 1 - \frac{1}{800}, \quad Y_i \geq 1 - \frac{1}{800}, \quad \frac{1}{\min_i(\|t_i - p_i\|_2^2)}Z_i \leq 0.1 n, \quad i = 1,2\ldots n \right\}
\]
On $E_2$, we have
\[
 \|h\|_2 \sum_{i=1}^n \sqrt{2}\left[ \sqrt{1 - X_i} + \sqrt{1 - Y_i} \right]  + \frac{1}{\min_i(\|t_i - p_i\|_2^2)}Z_i\|h\|_2 \leq 0.2 n \|h\|_2.
\]
We now estimate $\mathbb{P}(E_2)$. For $X_i$, since $ (t_i - p_i)$ is independent from $(x-y, p_i + t_i - x-y)$, we can view the latter as fixed. That is, by conditioning on $V_i$, and applying a rotation $R$ such  that $R(V_i) = \Span(e_1,e_2)$, we have
\[
 \frac{\| P_{V_i^\perp} (t_i-p_i) \|_2}{\|t_i-p_i\|_2} =^d \sqrt{\frac{\sum_{j=1}^{d-2}t_i (j)^2}{\sum_{j=1}^{d}t_i (j)^2}} 
\] 
where $t_i(j)$ is the $j$th entry of $t_i$.  As $\sum_{j=1}^{d-2}t_i (j)^2 \sim \chi^2_{d-2}$ and $\sum_{j=1}^{d}t_i (j)^2 \sim \chi^2_{d}$, Lemma \ref{lem:conc1} can be repeatedly applied to give $\P(X_i \geq 1 - \frac{1}{800} \text{ for all } i) \geq 1 - 2 n e^{-cd}$. 
A similar argument gives $\P(Y_i \geq 1 - \frac{1}{800} \text{ for all } i) \geq 1 - 2n e^{-cd}$ because  $x-y$ and $x+y-(t_i+p_i)$ are independent.

We now bound the probability of the third condition in the definition of $E_2$.  Note that 
\[
\frac{1}{\min_i(\|t_i - p_i\|_2^2)}\left \|\sum_{i=1}^n (t_i-p_i)(t_i-p_i)^*\right\|_{\text{op}} =^d \frac{1}{\min_i(\|t_i \|_2^2)}\left \|\sum_{i=1}^n t_i t_i^*\right\|_{\text{op}}
\]
Let $B = \sum_{i=1}^n e_i t_i^*$.  By Lemma \ref{lem:conc2}, $\| \sum_{i=1}^n t_i t_i^*\|_\text{op} = \sigma_\text{max}(B)^2 \geq n \Bigl(1 + 3 \sqrt{\frac{d}{n}} \Bigr)^2$ with probability at least $1 - 2 e^{-2d}$.
By Lemma \ref{lem:conc1}, $\|t_i\|_2^2 \geq d(1-\eps)$ for all $i$ with probability at least $1 - n e^{-c \eps^2 d}$.  We conclude
\[
 \frac{1}{\min_i(\|t_i \|_2^2)}\left \|\sum_{i=1}^n t_i t_i^*\right\|_{\text{op}} \leq \frac{n \Bigl(1 + 3 \sqrt{\frac{d}{n}} \Bigr)^2}{d (1-\eps)}
\]
with probability at least $1 - 2n e^{-c \eps^2 d}$.
If $\eps = 0.01, d \geq 40, n \geq 10d$, we have
\[
\P\Bigl( \frac{1}{\min_i(\|t_i - p_i\|_2^2)}Z_i \leq 0.1 n \Bigr)  \geq 1 - 2n e^{-cd}
\]
Hence, if $d \geq 40, n \geq 10 d$,
\[
\P(E_2) \geq 1 - 6n e^{-cd}.
\]

In conclusion, there exist positive integers $d_0$ and $n_0$ such that for all $d \geq d_0$, $n \geq n_0$, $n \geq 10 d$, and all $h \perp x-y$,
\[
\mathbb{P}\left(\left \| \sum_{i=1}^n P_{{W_i}^\perp} (h) \right\|_2 \geq \frac{1}{10}\|h\|_2 \right) \geq 1- \mathbb{P}\left[(E_1 \cap E_2)^c\right] \geq 1- 6ne^{-cd}
\]
for some $c > 0$, which implies the statement of the lemma.

\end{proof}



\subsection{Random graphs are $p$-typical with high probability} \label{sec-random-graph}

We prove that Condition 1 of Theorem~\ref{thm:mainthm} holds with high probability.

\begin{lemma} \label{lem:ptyp}
There exists an absolute constant $c > 0$ such that for all positive real numbers $p \le 1$ satisfying $n_2 p \ge 2\log (e n_1)$ and $n_1 p \ge 2\log (e n_2)$, $G(n_1,n_2;p)$ is $p$-typical with probability at least $1 - n_1 n_2 2^{n_1 + n_2} e^{-pn_1n_2/4} - n_1^2 n_2 e^{-\Omega(n_2p^2)} - n_1 n_2^2 e^{-\Omega(n_1p^2)}$.
\end{lemma}
\begin{proof}
Let $V_1$ and $V_2$ be vertex sets of sizes $|V_1| = n_1$ and $|V_2| = n_2$. 
Throughout the proof, we let $V_1 \cup V_2$ be the bipartition of the random graph $G(n_1,n_2;p)$.
The bipartite graph $G(n_1,n_2;p)$ is
not connected only if there exist partitions $V_1 = V_{1,1} \cup V_{1,2}$ 
and $V_2 = V_{2,1} \cup V_{2,2}$ such that the sets $V_{1,1} \cup V_{2,1}$
and $V_{1,2} \cup V_{2,2}$ are both non-empty and have no edges between them. 
Let $|V_{1,1}| = k_1, |V_{2,1}| = k_2$, $|V_{1,2}| = n_1 - k_1$ and $|V_{2,2}| = n_2 - k_2$. 
For fixed $k_1, k_2$, by the union bound, 
the probability that there exists a  partition as above is at most
\begin{align} \label{eq:disconnected}
	{n_1 \choose k_1} {n_2 \choose k_2} (1-p)^{k_1 (n_2 - k_2) + k_2 (n_1 - k_1)}.
\end{align}
If $k_1 \le \frac{n_1}{2}$ and $k_2 \le \frac{n_2}{2}$, then by Stirling's formula, \eqref{eq:disconnected}
is at most
\[
	\left(\frac{en_1}{k_1}\right)^{k_1} 
	\left(\frac{en_2}{k_2}\right)^{k_2}
	(1-p)^{(k_1 n_2 + k_2 n_1)/2}
	\le
	\left(\frac{en_1}{k_1} e^{-n_2p/2}\right)^{k_1} 
	\left(\frac{en_2}{k_2} e^{-n_1p/2}\right)^{k_2}.
\]
If $k_1 > \frac{n_1}{2}$ and $k_2 > \frac{n_2}{2}$, then 
let $\ell_1 = n_1 - k_1$ and $\ell_2 = n_2 - k_2$. Then \eqref{eq:disconnected} is at most
\[
	\left(\frac{en_1}{\ell_1}\right)^{\ell_1} 
	\left(\frac{en_2}{\ell_2}\right)^{\ell_2}
	(1-p)^{(\ell_1 n_2 + \ell_2 n_1)/2}
	\le
	\left(\frac{en_1}{\ell_1} e^{-n_2p/2}\right)^{\ell_1} 
	\left(\frac{en_2}{\ell_2} e^{-n_1p/2}\right)^{\ell_2}.
\]
If ($k_1 \le \frac{n_1}{2}$ and $k_2 > \frac{n_2}{2}$)
or ($k_1 > \frac{n_1}{2}$ and $k_2 \le \frac{n_2}{2}$), then, by ${n \choose k} \leq 2^n$ for all $0 \leq k \leq n$, \eqref{eq:disconnected}
is at most
\[
	2^{n_1 + n_2} (1-p)^{n_1 n_2/4}
	\le 2^{n_1 + n_2} e^{-pn_1n_2/4}.
\]

Hence the probability that $G(n_1, n_2;p)$ is disconnected is at most
\begin{align*}
	&\, \sum_{k_1=1}^{\lfloor n_1/2 \rfloor}
	\sum_{k_2=0}^{\lfloor n_2/2 \rfloor}
	\left(\frac{en_1}{k_1} e^{-n_2p/2}\right)^{k_1} 
	\left(\frac{en_2}{k_2} e^{-n_1p/2}\right)^{k_2}
	+
	\sum_{k_1=0}^{\lfloor n_1/2 \rfloor}
	\sum_{k_2=1}^{\lfloor n_2/2 \rfloor}
	\left(\frac{en_1}{k_1} e^{-n_2p/2}\right)^{k_1} 
	\left(\frac{en_2}{k_2} e^{-n_1p/2}\right)^{k_2}
	\\
	&\,+ \sum_{\ell_1=1}^{\lfloor n_1/2 \rfloor}
	\sum_{\ell_2=0}^{\lfloor n_2/2 \rfloor}
	\left(\frac{en_1}{\ell_1} e^{-n_2p/2}\right)^{\ell_1} 
	\left(\frac{en_2}{\ell_2} e^{-n_1p/2}\right)^{\ell_2}
	+
	\sum_{\ell_1=0}^{\lfloor n_1/2 \rfloor}
	\sum_{\ell_2=1}^{\lfloor n_2/2 \rfloor}
	\left(\frac{en_1}{\ell_1} e^{-n_2p/2}\right)^{\ell_1} 
	\left(\frac{en_2}{\ell_2} e^{-n_1p/2}\right)^{\ell_2}
	\\
	&\,+ n_1 n_2 2^{n_1 + n_2} e^{-pn_1n_2/4},
\end{align*}
where the indeterminate factors in the sums corresponding to $k_1=0$, $k_2=0$, $\ell_1=0$, or $\ell_2=0$ are taken to be unity.
Since $n_2 p \ge 2\log (e n_1)$ and $n_1 p \ge 2\log (e n_2)$, the four sums above are maximized
at $(k_1, k_2)=(1,0), (0,1)$, $(\ell_1, \ell_2)=(1,0), (0,1)$, respectively. Therefore
the probability that $G(n_1, n_2;p)$ is disconnected is at most
\begin{align*}
	&\, 
	2n_1 n_2
	\cdot en_1 \cdot e^{-n_2p/2}
	+
	2n_1 n_2
	\cdot en_2 \cdot e^{-n_1p/2}
	+ n_1 n_2 2^{n_1 + n_2} e^{-pn_1n_2/4}. 
\end{align*}

For a fixed vertex $v \in V_1$, the expected value of $\deg(v)$ is $n_2 p$, 
and for a pair of vertices $v,w \in V_1$, the expected value of
the codegree of $v$ and $w$ is $n_2 p^{2}$. Therefore by 
Chernoff's inequality (see Fact 4 from  \cite{AngluinValiant}) and a union bound, 
the probability that all vertices in $V_1$ have degree between $\frac{1}{2}n_2 p$ and $2n_2p$, 
and all pairs of vertices in $V_1$ have codegree between $\frac{1}{2} n_2p^2$ and $2n_2p^2$ is
$1 - n_1^2 e^{-\Omega(n_2p^2)}$.
Similarly, the probability that all vertices in $V_2$ have degree between $\frac{1}{2}n_1 p$ and $2n_1p$, 
and all pairs of vertices in $V_2$ have codegree between $\frac{1}{2} n_1p^2$ and $2n_1p^2$ is
$1 - n_2^2 e^{-\Omega(n_1p^2)}$. The conclusion follows by taking a union bound over
all events.
\end{proof}

\subsection{Proof of Theorem \ref{thm:bipartite-random}} \label{sec-proof-random-theorem}
We can now prove the high dimensional recovery theorem, which we state here again for convenience: 
\addtocounter{theorem}{-2}
\begin{theorem} 
Let $N = \max(\nl,\ns), n = \min(\nl, \ns)$. Let $G(\Vl \cup \Vs,E)$ be drawn from a bipartite-\erdosrenyi graph with $p >0$. Take $ \tnot_1, \ldots \tnot_{\nl}, \pnot_1, \ldots, \pnot_{\ns} \sim \mathcal{N}(0, I_{d \times d})$ to be independent from each other and $G$. Then, there exist absolute constants $c,c_3, C>0$ such that for $\gamma = c_3 p^4$, if \\
$$\max\left(\frac{1}{c_3 p^4}, Cd, \frac{2 \log(eN)}{p}, \Omega(c_3 \log^2 N)\right) \leq  n \le N \leq e^{\frac{1}{8}c d }$$ 
and $d = \Omega(1)$, then there exists an event with probability at least  $1 - O( e^{-\Omega(\frac{1}{2}c_3^{-1/2} n^{1/2})} + e^{-\frac{1}{2} cd})$, on which the following holds:\\[1em]
For all subgraphs $\Eb$ satisfying $\max_{i\in [\nl]} \deg_b(\ti) \leq \gamma \ns$ and $\max_{j \in [\ns]} \deg_b(\pj) \leq \gamma \nl$ and all pairwise direction corruptions  $\vij \in \mathbb{S}^{d-1}$ for $ij \in \Eb$,  the convex program \eqref{shapefit} has a unique minimizer equal to $\left\{\alpha \{\toi -\tpbar \}_{i \in [\nl]}, \alpha \{\poi -\tpbar\}_{j \in [\ns]}  \right\}$ for some positive $\alpha$ and for $\tpbar = \frac{1}{\nl + \ns} \left(\sum_{i \in [\nl]} \toi + \sum_{j \in [\ns]} \poj \right)$. 
\end{theorem}
\begin{proof}
Let $c$ be minimum of the constants from Lemmas \ref{lem-conds34} and \ref{lem-c4-well-distributed}. Let $K_0$ be the constant from Lemma \ref{lem-c4-well-distributed}.
It is enough to verify that $G$, $T$ and $E_b$ in the assumption of the present theorem satisfy the deterministic conditions 1--6 in Theorem \ref{thm:mainthm}, with appropriate constants $p, \beta, c_0, \epsilon, c_1$, and with the purported probability.
By Lemma \ref{lem:ptyp}, Condition 1 holds with probability at least 
$$1 - n_l n_s 2^{n_l + n_s} e^{-pn_ln_s/4} - n_l^2 n_s e^{-\Omega(n_sp^2)} - n_l n_s^2 e^{-\Omega(n_lp^2)} = 1 - O(N^3 e^{-\Omega(np^2)})$$
 if $np \geq 2 \log(eN)$.  Condition 2 holds with probability $1$. By Lemma \ref{lem-conds34}, Condition 3 holds for $c_0 = \frac{9}{10}$ with probability at least $1 - 2 \nl \ns e^{-cd}$, and Condition 4 holds for $\beta = \frac{1}{4}$ with probability at least $1 - 22 \nl^2 \ns^2 e^{-c d}$. 
By Lemma \ref{lem-c4-well-distributed}, Condition 5 holds for $c_1 = \frac{1}{20}$ with probability $1 - O( \nl^2 \ns^2 e^{-cd})$ if $n \geq \max(K_0, 160d) $, and $d \geq d_0$.
 Thus, Conditions 1--5 hold together with probability at least
\[
1 -  O(N^4 e^{-cd} +N^3 e^{-\Omega(n p^2)}).
\]

Take $\gamma = c_3 p^4 \leq \frac{p^4}{10^{11}}$.  Because $\gamma \leq \frac{\beta c_0 c_1^2 p^4}{384\cdot204\cdot64}$, Theorem \ref{thm:mainthm} implies that recovery via ShapeFit is guaranteed.  Note that the conditions $\max_{i\in\Vl} \deg_b(i) \leq \gamma \ns$ and $\max_{j \in \Vs} \deg_b(j) \leq \gamma \nl$ are nontrivial when $p \geq c_3^{-1/4} n^{-1/4}$.  Using this inequality, we have $N^3 e^{-\Omega(np^2)} \leq N^3 e^{-\Omega(c_3^{-1/2} n^{1/2})} \leq e^{-\Omega(\frac{1}{2} c_3^{-1/2} n^{1/2})}$ if $n = \Omega(c_3 \log^2 N) $ and $N^4 e^{-cd} \le e^{-cd/2}$ if $N \leq e^{\frac{1}{8}c d}$. Thus, the probability of exact recovery via ShapeFit, uniformly in $E_b$ and $v_{ij}$ satisfying the assumptions of the theorem, is at least
\[
1- O(e^{-\Omega(\frac{1}{2}c_3^{-1/2} n^{1/2})} + e^{-\frac{1}{2}c d}). \qedhere
\]


\end{proof}

\section{Numerical simulations} \label{sec-simulations}

In this section, we use numerical simulation to verify that ShapeFit recovers Gaussian camera locations and Gaussian structure locations in $\R^3$ in the presence of corrupted pairwise direction measurements.  Further, we empirically demonstrate that ShapeFit is robust to noise in the uncorrupted measurements.

Let $\toitilde \in \R^3$ be independent $\mathcal{N}(0, I_{3 \times 3})$ random variables for $i = 1, \ldots, \nl$.  Let $\pojtilde \in \R^3$ be independent $\mathcal{N}(0, I_{3 \times 3})$ random variables for $j = 1, \ldots, \ns$.  
Let 
\[\toi = \toitilde - \frac{1}{\nl + \ns}\Bigl( \sum_k \toktilde + \sum_\ell \poltilde \Bigr) \text{ and } \poj = \pojtilde -  \frac{1}{\nl + \ns} \Bigl( \sum_k \toktilde + \sum_\ell \poltilde \Bigr).
\]
Let the graph of observations $G$ be a bipartite \erdosrenyi graph  $G(\nl, \ns, p)$ on $\nl + \ns$ vertices, for $p = 1/2$.
  For $ij \in E(G)$, let
\[
\vtildeij = \begin{cases} \zij & \text{ with probability } q,  \\[0.2em]
\frac{\toi - \poj}{\| \toi - \poj\|_2} + \sigma \zij & \text{ with probability } 1-q,
\end{cases}
\]
where $\zij$ are independent and uniform over $\mathbb{S}^2$. Let $\vij = \vtildeij / \| \vtildeij\|_2$. That is, each observation is corrupted with probability $q$, and each corruption is in a random direction.  In the noiseless case, with $\sigma = 0$, each observation is exact with probability $1-q$.  

We solved ShapeFit using the SDPT3 solver \cite{TTT1999, TTT2003} and YALMIP \cite{L2004}.  For output $S = (T,P) = \Bigl(\{ t_i\}_{i \in [\nl]}, \{ p_j\}_{j \in [\ns]}\Bigr)$, define its relative error with respect to $\Snot = (\Tnot, \Pnot) =$ $\Bigl(\{\toi\}_{i \in [\nl]},\{\poj\}_{j \in [\ns]} \Bigr)$ as
\[
\left \| \frac{S}{\|S\|_F} - \frac{\Snot}{\|\Snot\|_F} \right \|_F 
\]
where $\|S\|_F$ is the Frobenius norm of the matrix whose column are given by $\{t_i\}$ and $\{p_j\}$.    This error metric amounts to an $\ell_2$ norm after rescaling.

\begin{figure} 
\begin{center}
{\large
%
%
\begin{psfrags}%
\psfragscanon%
%
\psfrag{s07}[][]{\color[rgb]{0,0,0}\setlength{\tabcolsep}{0pt}\begin{tabular}{c}Number of locations ($\nl+\ns$)\end{tabular}}%
\psfrag{s08}[][]{\color[rgb]{0,0,0}\setlength{\tabcolsep}{0pt}\begin{tabular}{c}ShapeFit under corruptions and noise\end{tabular}}%
\psfrag{s09}[][]{\color[rgb]{0,0,0}\setlength{\tabcolsep}{0pt}\begin{tabular}{c}ShapeFit under corruptions and no noise\end{tabular}}%
\psfrag{s11}[][]{\color[rgb]{0,0,0}\setlength{\tabcolsep}{0pt}\begin{tabular}{c}Number of locations ($\nl+\ns$)\end{tabular}}%
\psfrag{s12}[][]{\color[rgb]{0,0,0}\setlength{\tabcolsep}{0pt}\begin{tabular}{c}Corruption probability ($q$)\end{tabular}}%
\psfrag{s13}[][]{\color[rgb]{0,0,0}\setlength{\tabcolsep}{0pt}\begin{tabular}{c}Corruption probability ($q$)\end{tabular}}%
%
\color[rgb]{0.15,0.15,0.15}%
%
\psfrag{x01}[t][t]{0}%
\psfrag{x02}[t][t]{0.1}%
\psfrag{x03}[t][t]{0.2}%
\psfrag{x04}[t][t]{0.3}%
\psfrag{x05}[t][t]{0.4}%
\psfrag{x06}[t][t]{0.5}%
\psfrag{x07}[t][t]{0}%
\psfrag{x08}[t][t]{0.1}%
\psfrag{x09}[t][t]{0.2}%
\psfrag{x10}[t][t]{0.3}%
\psfrag{x11}[t][t]{0.4}%
\psfrag{x12}[t][t]{0.5}%
%
\psfrag{v01}[r][r]{70}%
\psfrag{v02}[r][r]{50}%
\psfrag{v03}[r][r]{30}%
\psfrag{v04}[r][r]{10}%
\psfrag{v05}[r][r]{70}%
\psfrag{v06}[r][r]{50}%
\psfrag{v07}[r][r]{30}%
\psfrag{v08}[r][r]{10}%
%
\resizebox{14cm}{!}{\includegraphics{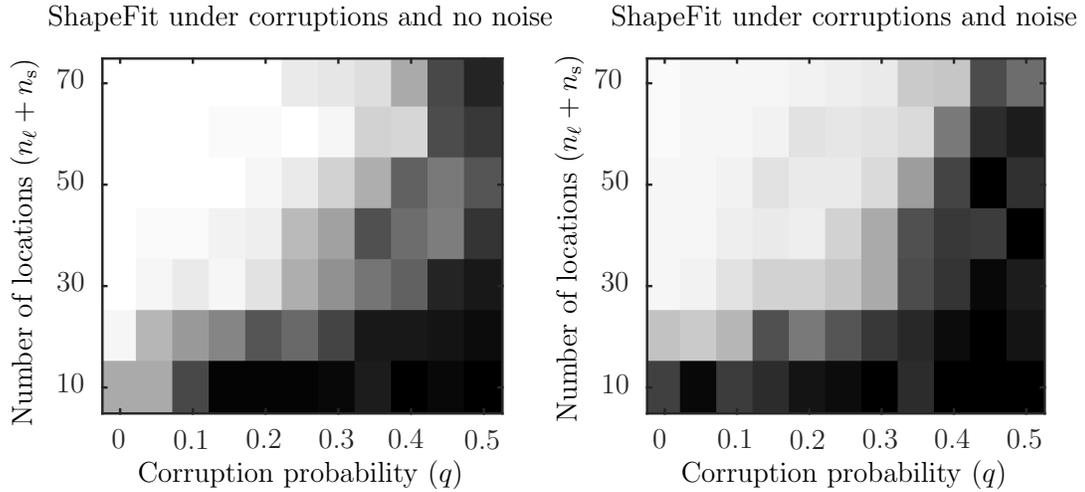}}%
\end{psfrags}%
%

}
\end{center}
\caption{Mean recovery error of ShapeFit as a function of the number of locations $\nl + \ns$ and the corruption probability $q$.  The data model has 3d Gaussian locations whose pairwise directions are observed in accordance with a bipartite \erdosrenyi graph $G(\nl, \ns, 1/2)$ and are corrupted with probability $q$.   White blocks represent an average relative error of zero over 10 independently generated problems.  Black blocks represent an average relative error of 100\%.   The left panel corresponds to the noiseless case $\sigma = 0$, and the right panel corresponds to the noisy case $\sigma = 0.05$.  
}
\label{fig-phase-transition}
\end{figure}

Figure \ref{fig-phase-transition} shows the mean relative error of the output of ShapeFit over 10 independent trials for locations in $\R^3$ generated by $p = 1/2$, $\nl = \ns$, $\sigma \in [0, 0.05]$, and a range of values $10 \leq \nl + \ns \leq 70$ and $0 \leq q \leq 0.5$.   White blocks represent zero average relative error, and black blocks represent an average relative error of 1 or higher.  Average residuals between $0$ and $1$ are represented by the appropriate shade of gray.  The figure shows that ShapeFit can empirically recover 3d locations in the presence of a surprisingly large probability of corruption, provided $n$ is big enough. For example, if $n \geq 50$, ShapeFit outputs a structure with small relative error even when around 15\% of all measurements are randomly corrupted.  Further, successful recovery occurs both in the noiseless case, and in the noisy case with $\sigma=0.05$.

\begin{figure}
\begin{center}
{\large
%
%
\begin{psfrags}%
\psfragscanon%
%
\psfrag{s02}[][]{\color[rgb]{0,0,0}\setlength{\tabcolsep}{0pt}\begin{tabular}{c}Noise parameter $\sigma$\end{tabular}}%
\psfrag{s03}[][]{\color[rgb]{0,0,0}\setlength{\tabcolsep}{0pt}\begin{tabular}{c}Median relative error\end{tabular}}%
\psfrag{s04}[][]{\color[rgb]{0,0,0}\setlength{\tabcolsep}{0pt}\begin{tabular}{c}ShapeFit for $\nl=\ns=25$, $q=0.1$\end{tabular}}%
%
\color[rgb]{0.15,0.15,0.15}%
%
\psfrag{x01}[t][t]{$10^{-6}$}%
\psfrag{x02}[t][t]{$10^{-5}$}%
\psfrag{x03}[t][t]{$10^{-4}$}%
\psfrag{x04}[t][t]{$10^{-3}$}%
\psfrag{x05}[t][t]{$10^{-2}$}%
\psfrag{x06}[t][t]{$10^{-1}$}%
\psfrag{x07}[t][t]{$10^{0}$}%
%
\psfrag{v01}[r][r]{$10^{-6} \hspace{-0.1em}$}%
\psfrag{v02}[r][r]{}%
\psfrag{v03}[r][r]{$10^{-4}$}%
\psfrag{v04}[r][r]{}%
\psfrag{v05}[r][r]{$10^{-2}$}%
\psfrag{v06}[r][r]{}%
\psfrag{v07}[r][r]{$10^{0}\ \hspace{0.1825em} $}%
%
\resizebox{9cm}{!}{\includegraphics{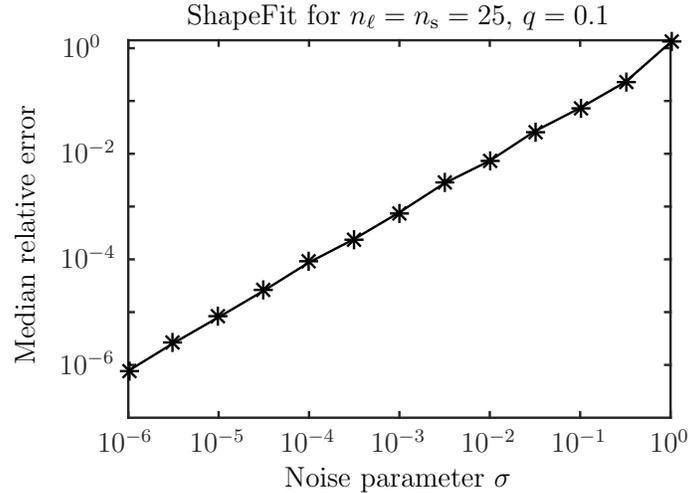}}%
\end{psfrags}%
%

}
\end{center}
\caption{
Median recovery error of ShapeFit versus the noise parameter $\sigma$.  These simulations are based on $50$  Gaussian locations in $\R^3$ whose pairwise directions are observed in accordance with a bipartite \erdosrenyi graph $G(25, 25,1/2)$ and are corrupted with probability $q=0.1$.  The median is based on 10 independently generated problems.
}
 \label{fig-robustness}
\end{figure}

Figure \ref{fig-robustness} shows the median residual over 10 independent trials for locations in $\R^3$ generated by $p = 1/2$, $\nl = \ns = 25$, $q= 0.1$ and a range of values of $10^{-6} \leq \sigma \leq 10^0$.  We see that ShapeFit is empirically stable to noise, with median residuals that are approximately linear in the noise parameter $\sigma$.
 
\section*{Acknowledgements}
We are grateful to Stefano Soatto for suggesting to VV the problem formulation addressed in this paper. VV is partially supported by the Office of Naval Research.  CL is partially supported by the National Science Foundation Grant DMS-1362326. PH is partially supported by the National Science Foundation Grant DMS-1418971.
\bibliographystyle{plain}
\bibliography{references}

\end{document}